\documentclass{article}

\usepackage{times}

\usepackage{graphicx} 
\usepackage{subfigure}

\usepackage{amsmath}
\usepackage{mathrsfs}
\usepackage{amsfonts}
\usepackage{amsthm}
\usepackage{stmaryrd}

\usepackage{natbib}

\usepackage{algorithm}
\usepackage{algorithmic}

\usepackage{hyperref}

\usepackage[accepted]{icml2017}

\newcommand{\noappendix}[1]{#1}

\newcommand{\gamename}[1]{\textsc{#1}}

\usepackage{enumitem}

\newcommand{\cX}{\mathcal{X}}

\newcommand{\figref}[1]{{Fig.\ \ref{fig:#1}}}

\DeclareMathOperator{\pg}{PG}
\DeclareMathOperator{\rc}{N}
\DeclareMathOperator{\pc}{\hat N}

\icmltitlerunning{Count-Based Exploration with Neural Density Models}

\begin{document}

\twocolumn[
\icmltitle{Count-Based Exploration with Neural Density Models}

\begin{icmlauthorlist}
\icmlauthor{Georg Ostrovski}{dm}
\icmlauthor{Marc G. Bellemare}{dm}
\icmlauthor{A{\"a}ron van den Oord}{dm}
\icmlauthor{R{\'e}mi Munos}{dm}
\end{icmlauthorlist}

\icmlaffiliation{dm}{DeepMind, London, UK}
\icmlcorrespondingauthor{Georg Ostrovski}{ostrovski@google.com}


\vskip 0.3in
]
\printAffiliationsAndNotice{}

\begin{abstract}
\citet{bellemare16cts} introduced the notion of a pseudo-count, derived from a density model, to generalize count-based exploration to non-tabular reinforcement learning.
This pseudo-count was used to generate an exploration bonus for a DQN agent and combined with a mixed Monte Carlo update was sufficient to achieve state of the art on the Atari 2600 game Montezuma's Revenge.
We consider two questions left open by their work: First, how important is the quality of the density model for exploration? Second, what role does the Monte Carlo update play in exploration? We answer the first question by demonstrating the use of PixelCNN, an advanced neural density model for images, to supply a pseudo-count. In particular, we examine the intrinsic difficulties in adapting \citeauthor{bellemare16cts}'s approach when assumptions about the model are violated. The result is a more practical and general algorithm requiring no special apparatus. We combine PixelCNN pseudo-counts with different agent architectures to dramatically improve the state of the art on several hard Atari games. One surprising finding is that the mixed Monte Carlo update is a powerful facilitator of exploration in the sparsest of settings, including Montezuma's Revenge.
\end{abstract}

\section{Introduction}

Exploration is the process by which an agent learns about its environment. In the reinforcement learning framework, this involves reducing the agent's uncertainty
about the environment's transition dynamics and attainable rewards. From a theoretical perspective, exploration is now well-understood \citep[e.g.][]{strehl08analysis,jaksch10nearoptimal,osband16generalization}, and Bayesian methods have been successfully demonstrated in a number of settings \citep{deisenroth11pilco,guez12efficient}. On the other hand, practical algorithms for the general case remain scarce; fully Bayesian approaches are usually intractable in large state spaces, and the count-based method typical of theoretical results is not applicable in the presence of value function approximation.

Recently, \citet{bellemare16cts} proposed the notion of \emph{pseudo-count} as a reasonable generalization of the tabular setting considered in the theory literature.
The pseudo-count is defined in terms of a density model $\rho$ trained on the sequence of states experienced by an agent:
\begin{equation*}
    \pc(x) = \rho(x) \hat n(x),
\end{equation*}
where $\hat n(x)$ can be thought of as a total pseudo-count computed from the model's \emph{recoding probability} $\rho'(x)$, the probability of $x$ computed immediately after training on $x$. As a practical application the authors used the pseudo-counts derived from the simple CTS density model \citep{bellemare14cts} to incentivize exploration in Atari 2600 agents. One of the main outcomes of their work was substantial empirical progress on the infamously hard game \textsc{Montezuma's Revenge}.

Their method critically hinged on several assumptions regarding the density model: 1) the model should be \emph{learning-positive}, i.e.\ the probability assigned to a state $x$ should increase with training; 2) it should be trained online, using each sample exactly once; and 3) the effective model step-size should decay at a rate of $n^{-1}$. Part of their empirical success also relied on a mixed Monte Carlo/Q-Learning update rule, which permitted fast propagation of the exploration bonuses.

In this paper, we set out to answer several research questions related to these modelling choices and assumptions:
\begin{enumerate}
 \item To what extent does a better density model give rise to better exploration?

 \item Can the above modelling assumptions be relaxed without sacrificing exploration performance?

 \item What role does the mixed Monte Carlo update play in successfully incentivizing exploration?

\end{enumerate}

In particular, we explore the
use of PixelCNN \cite{oord2016pixela, oord2016pixelb},
a state-of-the-art neural density model. We examine the challenges posed by this approach:

\textbf{Model choice.}
Performing two evaluations and one model update \textit{at each agent step} (to compute $\rho(x)$ and $\rho'(x)$) can be
prohibitively expensive. This
requires the design of a simplified -- yet sufficiently expressive and accurate -- PixelCNN architecture.

\textbf{Model training.}
A CTS model can naturally be
trained from sequentially presented, correlated data samples.
Training a \textit{neural} model in this online fashion requires more careful
attention to the optimization procedure
to prevent overfitting and catastrophic forgetting \cite{french1999catastrophic}.

\textbf{Model use.}
The theory of pseudo-counts requires the density model's rate of learning
to decay over time. Optimization of a neural model,
however, imposes constraints on the step-size regime
which cannot be violated without deteriorating effectiveness and stability of training.

The concept of intrinsic motivation has made a recent resurgence in reinforcement learning research, in great part due to a dissatisfaction with $\epsilon$-greedy and Boltzmann policies. Of note, \citet{tang16exploration} maintain an approximate count by means of hash tables over features, which in the pseudo-count framework corresponds to a hash-based density model. \citet{houthooft16curiosity} used a second-order Taylor approximation of the prediction gain to drive exploration in continuous control.
As research moves towards ever more complex environments, we expect the trend towards more intrinsically motivated solutions to continue.

\section{Background}

\subsection{Pseudo-Count and Prediction Gain}

Here we briefly introduce notation and results, referring the reader to
\cite{bellemare16cts} for technical details.

Let $\rho$ be a density model on a finite space $\cX$, and $\rho_n(x)$ the probability assigned by
the model to $x$ after being trained on a sequence of states $x_1, \ldots, x_n$. Assume $\rho_n(x) > 0$ for all $x, n$.
The \emph{recoding probability} $\rho'_n(x)$ is then the probability the model would assign to $x$ if it were trained on that same $x$ one more time.
We call $\rho$ \textit{learning-positive} if
$\rho_n'(x) \geq \rho_n(x)$ for all $x_1, \ldots, x_n, x \in \cX$.
The \textit{prediction gain} (PG) of $\rho$ is
\begin{equation}
\pg_n(x) = \log \rho'_n(x) - \log \rho_n(x).
\end{equation}
A learning-positive $\rho$ implies $\pg_n(x) \geq 0$ for all $x \in \cX$.
For learning-positive $\rho$, we define the \textit{pseudo-count} as
\begin{equation*}
\pc_n(x) = \frac{\rho_n(x)(1-\rho_n'(x))}{\rho_n'(x)-\rho_n(x)},
\end{equation*}
derived from postulating that a single observation of $x \in \cX$
should lead to a unit increase in pseudo-count:
\begin{equation*}
\rho_n(x) = \frac{\pc_n(x)}{\hat n}, \quad \rho'_n(x) = \frac{\pc_n(x) + 1}{\hat n + 1},
\end{equation*}
where $\hat n$ is the \textit{pseudo-count total}.
The pseudo-count generalizes the usual state visitation count function $\rc_n(x)$.
Under certain assumptions on $\rho_n$,
pseudo-counts grow approximately linearly with real counts. Crucially,
the pseudo-count can be approximated using the prediction gain of the
density model:
\begin{equation*}
\pc_n(x) \approx \left(e^{\pg_n(x)} - 1\right)^{-1}.
\end{equation*}
Its main use is to define an \emph{exploration bonus}.
We consider a reinforcement learning (RL) agent interacting with
an environment that provides observations and extrinsic rewards
\citep[see][for a thorough exposition of the RL framework]{sutton98reinforcement}.
To the reward at step $n$ we add the bonus
\begin{equation*}
    r^+(x) := (\pc_n(x))^{-1/2},
\end{equation*}
which incentivizes the agent to try to re-experience surprising situations.
Quantities related to prediction gain have been used for similar purposes
in the intrinsic motivation literature \citep{lopes12exploration},
where they measure an agent's \emph{learning progress} \citep{oudeyer07intrinsic}.
Although the pseudo-count bonus is close to the prediction gain,
it is asymptotically more conservative and supported by stronger theoretical guarantees.

\subsection{Density Models for Images}

\begin{figure}
\begin{center}
\includegraphics[height=1.2in]{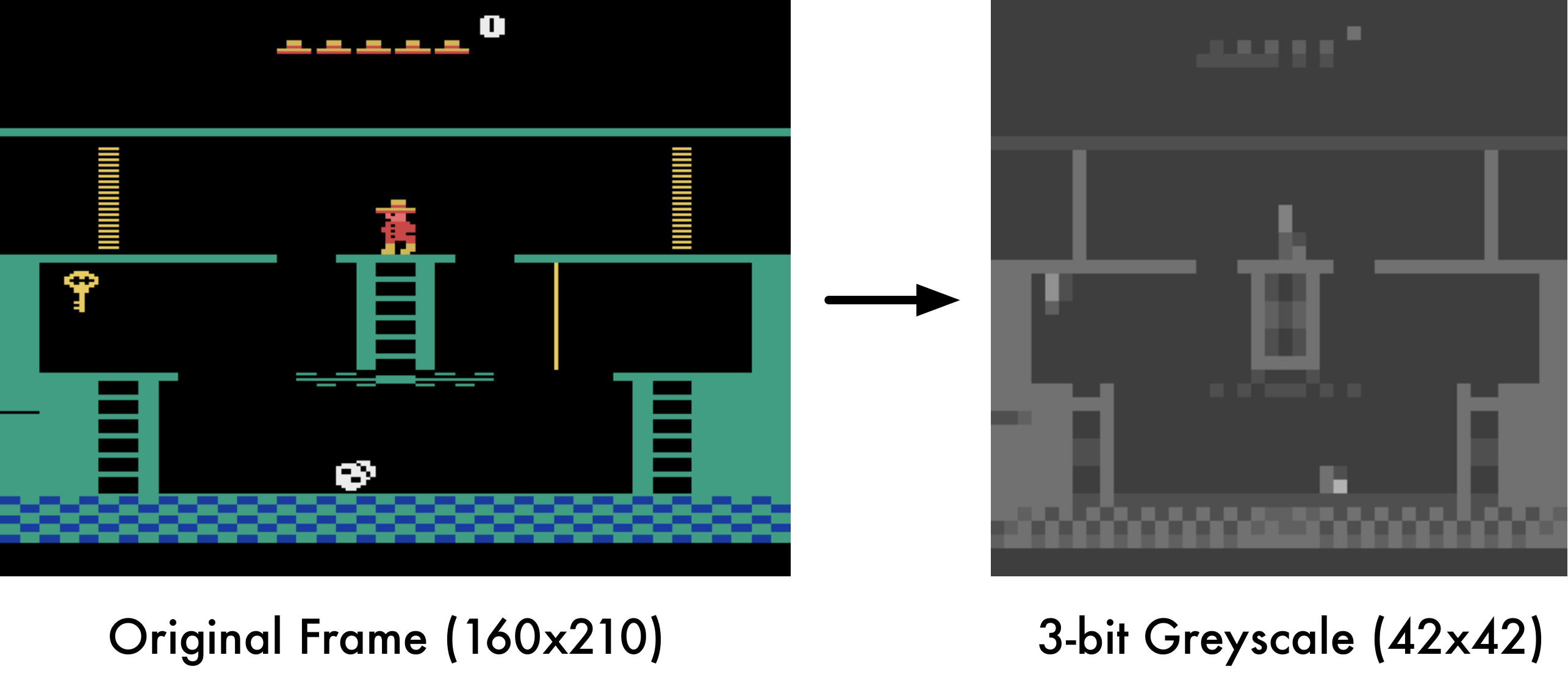}
\vspace{-1em}
\caption{Atari frame preprocessing  \citep{bellemare16cts}.}
\label{fig:cts_downsampling}
\end{center}
\end{figure}

The CTS density model \citep{bellemare14cts} is based on the namesake algorithm, Context Tree Switching \citep{veness12context},
a Bayesian variable-order Markov model.
In its simplest form, the model takes as input a 2D image and assigns to it
a probability according to the product of location-dependent L-shaped filters,
where the prediction of each filter is given by a CTS algorithm trained on past images.
In \citet{bellemare16cts}, this model was applied to 3-bit greyscale, $42 \times 42$ downsampled Atari 2600 frames (\figref{cts_downsampling}).
The CTS model presents advantages in terms of simplicity and performance but is limited in expressiveness, scalability, and data efficiency.

In recent years, neural generative models for images have achieved impressive
successes in their ability to generate diverse images in various domains
\cite{kingma2013auto,rezende2014stochastic,gregor2015draw,goodfellow2014generative}.
In particular, \citet{oord2016pixela, oord2016pixelb}
introduced \emph{PixelCNN}, a fully convolutional neural network
composed of residual blocks with multiplicative gating units,
which models pixel probabilities conditional on previous pixels
(in the usual top-left to bottom-right raster-scan order)
by using masked convolution filters.
This model achieved state-of-the-art modelling performance on
standard datasets, paired with the
computational efficiency of a convolutional feed-forward network.

\subsection{Multi-Step RL Methods}

A distinguishing feature of reinforcement learning is that the agent ``learns on the basis of interim estimates'' \citep{sutton96generalization}. For example, the Q-Learning update rule is
\begin{align*}
    Q(x,a) &\gets Q(x,a) \\
    &+ \alpha \underbrace{\left [ r(x,a) + \gamma \max\nolimits_{a'} Q(x', a') - Q(x,a) \right ]}_{\delta(x,a)},
\end{align*}
linking the reward $r$ and next-state value function $Q(x', a')$ to the current state value function $Q(x,a)$. This particular form is the stochastic update rule with step-size $\alpha$ and involves the \emph{TD-error} $\delta$.
In the approximate reinforcement learning setting, such as when $Q(x,a)$ is represented by a neural network, this update is converted into a loss to be minimized, most commonly the squared loss $\delta^2(x,a)$.

It is well known that better performance, both in terms of learning efficiency and approximation error, is attained by multi-step methods \citep{sutton96generalization,tsitsiklis97analysis}. These methods interpolate between one-step methods (Q-Learning) and the Monte-Carlo update
\begin{equation*}
    Q(x,a) \gets Q(x,a) + \alpha \underbrace{\left [ \sum_{t=0}^\infty \gamma^t r(x_t, a_t) - Q(x,a) \right ]}_{\delta_{\textsc{mc}}(x,a)},
\end{equation*}
where $x_0, a_0, x_1, a_1, \dots$ is a sample path through the environment beginning in $(x,a)$.
To achieve their success on the hardest Atari 2600 games, \citet{bellemare16cts} used the \emph{mixed Monte-Carlo update} (MMC)
\begin{equation*}
    Q(x,a) \gets Q(x,a) + \alpha \left [ (1 - \beta) \delta(x,a) + \beta \delta_{\textsc{mc}}(x,a) \right ],
\end{equation*}
with $\beta \in [0, 1]$.
This choice was made for ``computational and implementational simplicity'', and is a particularly coarse multi-step method.
A better multi-step method is the recent Retrace($\lambda$) algorithm \citep{munos2016safe}. Retrace($\lambda$) uses a product of truncated importance sampling ratios $c_1, c_2, \dots$ to replace $\delta$ with the error term
\begin{equation*}
    \delta_{\textsc{Retrace}}(x,a) := \sum_{t=0}^\infty \gamma^t \left(\prod_{s=1}^t c_s \right) \delta(x_t, a_t),
\end{equation*}
effectively mixing in TD-errors from all future time steps.
\citeauthor{munos2016safe} showed that Retrace$(\lambda)$ is safe
(does not diverge when trained on data from an arbitrary behaviour policy),
and efficient (makes the most of multi-step returns).

\section{Using PixelCNN for Exploration}

\begin{figure*}[tb]
\center{
\includegraphics[width=1.5in]{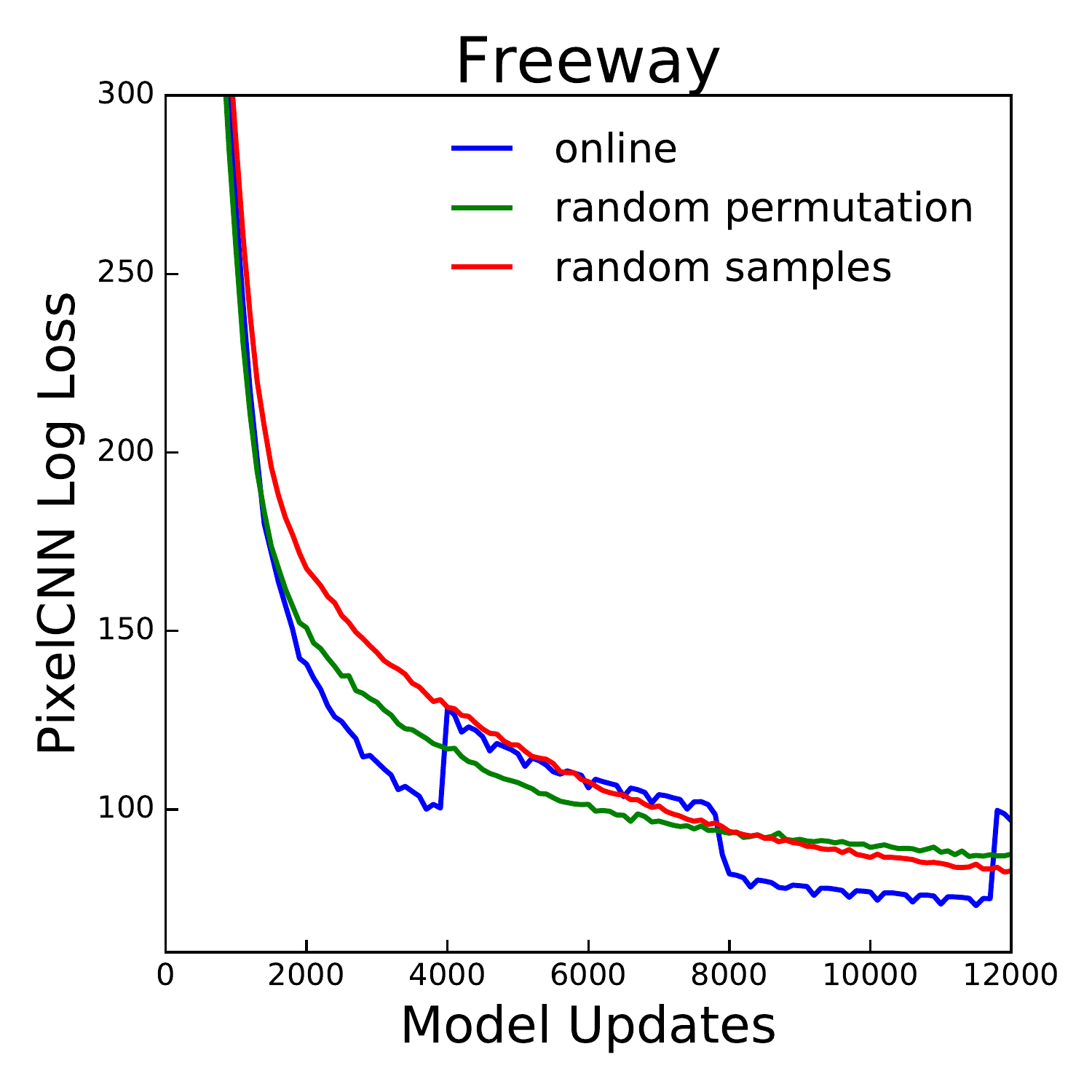}
\includegraphics[width=2.4in]{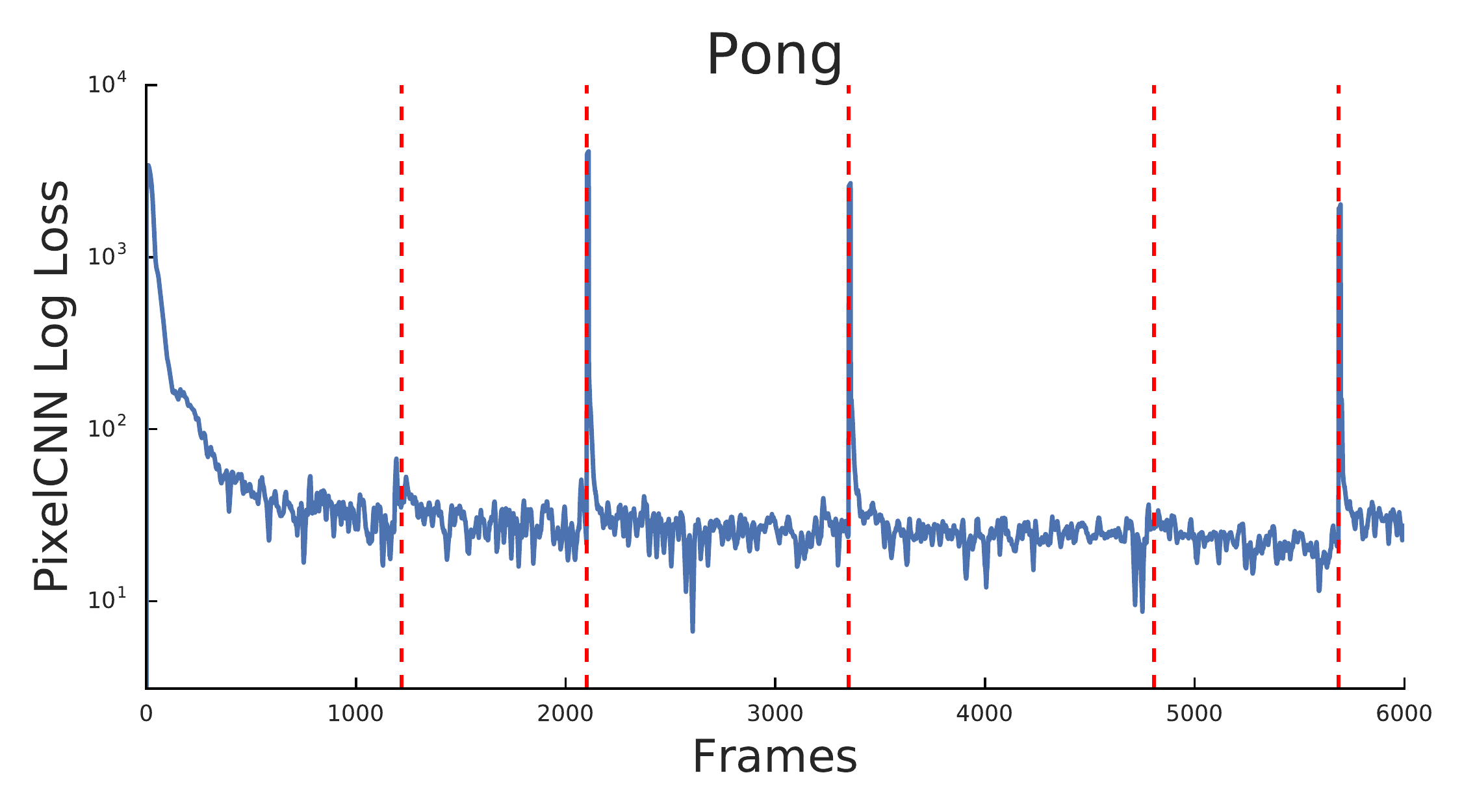}
\includegraphics[width=2.6in]{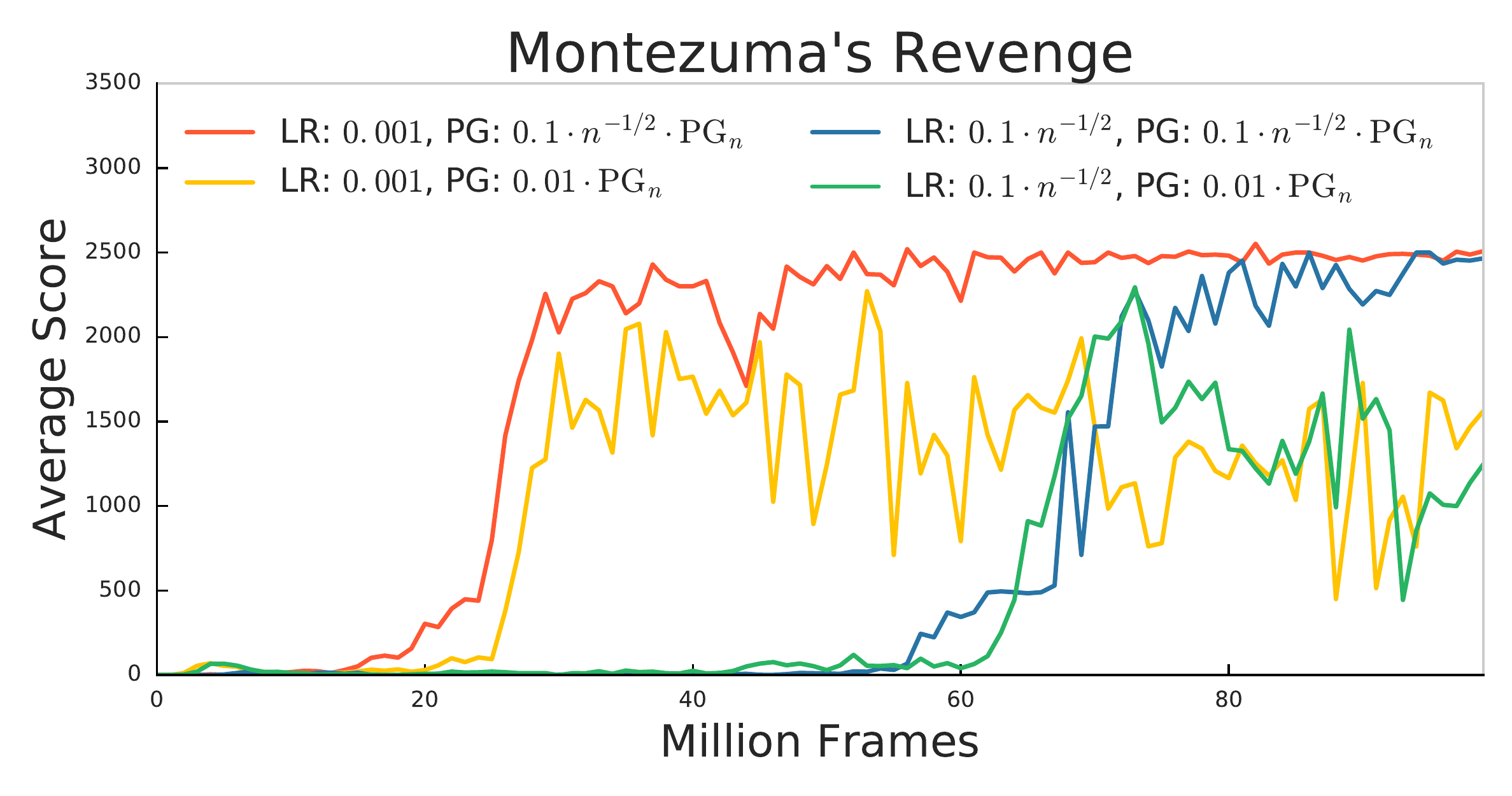}
}
\vspace{-1em}
\caption{\textbf{Left:} PixelCNN log loss on
\textsc{Freeway}, when tra\textbf{}ined \textit{online},
on a \textit{random permutation} (single use of each frame)
or on \textit{randomly drawn samples} (with replacement, potentially using same frame multiple times) from the state sequence.
To simulate the effect of non-stationarity, the agent's policy changes every 4K updates.
All training methods
show qualitatively similar learning progress and stability.
\textbf{Middle:} PixelCNN log loss over first 6K
training frames on \textsc{Pong}.
Vertical dashed lines indicate episode ends. The coinciding loss spikes
are the density model's `surprise' upon observing the distinctive green frame that sometimes occurs at the episode start.
\textbf{Right:} DQN-PixelCNN training performance on
\textsc{Montezuma's Revenge} as we vary learning rate and PG decay schedules.}
\label{fig:optim}
\end{figure*}

\begin{figure}[tb]
\center{
\includegraphics[width=3.3in,trim={0 0 70 0},clip]{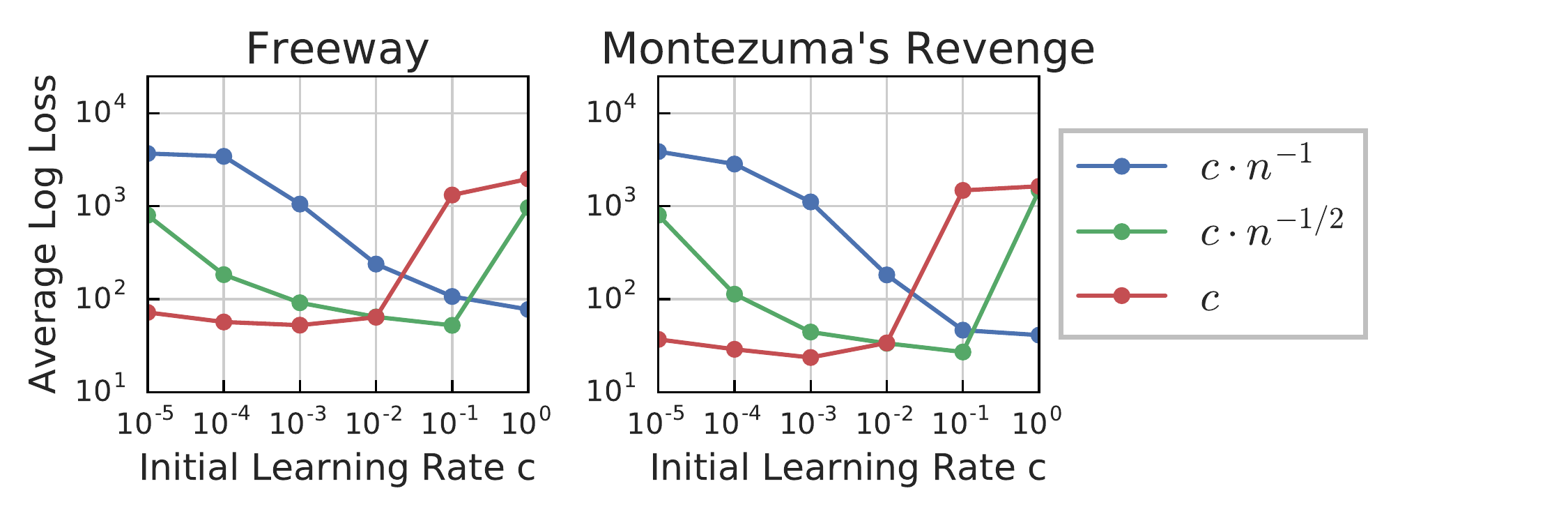}
}
\vspace{-2em}\caption{Model loss averaged over 10K frames, after 1M training frames, for constant, $n^{-1}$, and $n^{-1/2}$ learning rate schedules.
The smallest loss
is achieved by a constant learning rate of $10^{-3}$.}\label{fig:loss_conv}
\end{figure}

As mentioned in the Introduction,
the theory of using density models for exploration makes several assumptions that
translate into concrete requirements for an implementation:
\begin{enumerate}[label={(\alph*)}]
 \item The density model should be trained completely online, i.e.~exactly
 once on each state experienced by the agent, in the given sequential order.
 \item The prediction gain (PG) should decay at a rate $n^{-1}$ to ensure that
 pseudo-counts grow approximately linearly with real counts.
 \item The density model should be learning-positive.
\end{enumerate}

Simultaneously, a partly competing set of requirements are posed by the practicalities
of training a neural density model and using it as part of an RL agent:
\begin{enumerate}[label={(\alph*)}]
 \setcounter{enumi}{3}
 \item For stability, efficiency, and to avoid catastrophic forgetting
 in the context of a drifting data distribution, it is advantageous
 to train a neural model in mini-batches, drawn randomly
 from a diverse dataset.
 \item For effective training, a certain optimization regime
 (e.g.~a fixed learning rate schedule) has to be followed.
 \item The density model must be computationally lightweight,
 to allow computing the PG (two model evaluations and
 one update) as part of every training step of an RL agent.
\end{enumerate}

We investigate how to best resolve these tensions in the context of the Arcade Learning Environment \citep{bellemare2013arcade}, a suite of benchmark Atari 2600 games.

\subsection{Designing a Suitable Density Model}

Driven by (f) and aiming
for an agent with computational performance comparable to DQN,
we design a slim variant of the PixelCNN network.
Its core is a stack of 2 gated residual blocks with 16 feature maps
(compared to 15 residual blocks with 128 feature maps
in vanilla PixelCNN).
As was done with the CTS model, images are downsampled to
$42 \times 42$ and quantized to 3-bit greyscale.
See Appendix \ref{sec:hyperparams} for technical details.

\subsection{Training the Density Model}

Instead of using randomized mini-batches, we train the density model completely
online on the sequence of experienced states.
Empirically we found that with minor tuning of optimization
hyper-parameters
we could train the model as robustly on a temporally correlated
sequence of states as on a sequence with randomized order (\figref{optim}(left)).

Besides satisfying the theoretical requirement (a),
completely online training of the
density model has the advantage
that $\rho'_n = \rho_{n+1}$, so that the model update performed for
computing the PG need not be reverted\footnote{The CTS model allows querying the PG cheaply,
without incurring an actual update of model parameters.}.

Another more subtle reason for avoiding mini-batch updates of the density model
(despite (d)) is a practical optimization issue.
The (necessarily online) computation of the
PG involves a model update and hence the use of an optimizer.
Advanced optimizers used with deep neural networks, like the RMSProp optimizer
\cite{tieleman2012lecture} used in this work, are stateful,
tracking running averages of e.g.\ mean and variance of the model parameters.
If the model is \textit{additionally} trained from mini-batches,
the two streams of updates may
show different statistical characteristics (e.g.\ different gradient magnitudes),
invalidating the assumptions underlying the optimization algorithm
and leading to slower or unstable training.

To determine a suitable online learning rate schedule,
we train the model on a sequence of 1M frames of experience of a
random-policy agent.
We compare the loss achieved
by training procedures following constant or decaying learning rate
schedules, see \figref{loss_conv}.
The lowest final training loss
is achieved by a constant learning rate of $0.001$
or a decaying learning rate of $0.1 \cdot n^{-1/2}$. We settled our choice on the constant learning rate schedule as it showed greater robustness
with respect to the choice of initial learning rate.

PixelCNN rapidly learns a sensible distribution over state
space. \figref{optim}(left) shows the model's loss decaying as it
learns to exploit image regularities. Spikes in its loss function
quickly start to correspond to visually meaningful events, such as
the starts of episodes (\figref{optim}(middle)). A video of early
density model training is provided in \url{http://youtu.be/T6iaa8Z4eyE}.

\begin{figure}[htb]
\begin{center}
\includegraphics[width=1.3in,trim={120 50 100 50},clip]{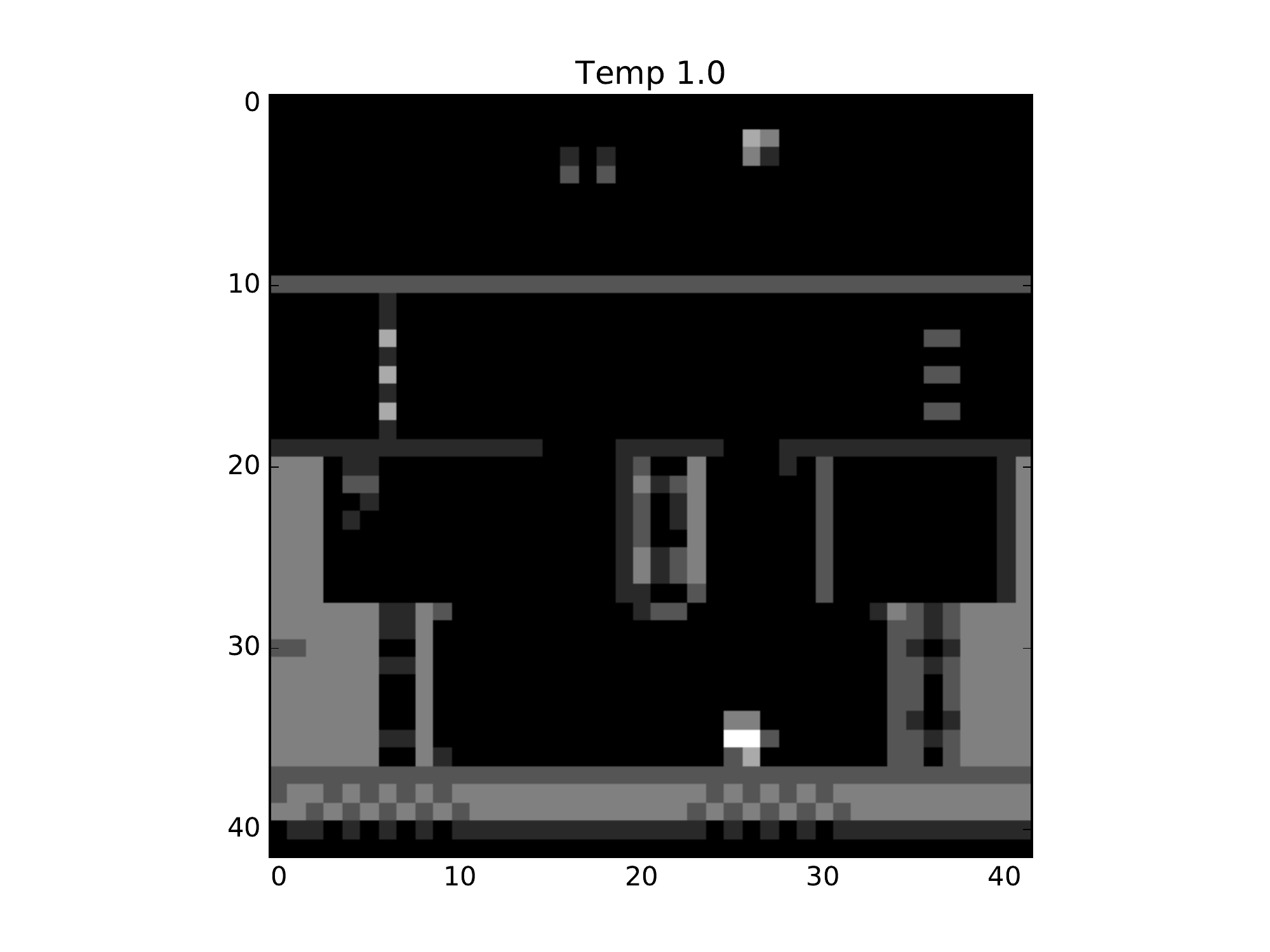}
\includegraphics[width=1.3in,trim={120 50 100 50},clip]{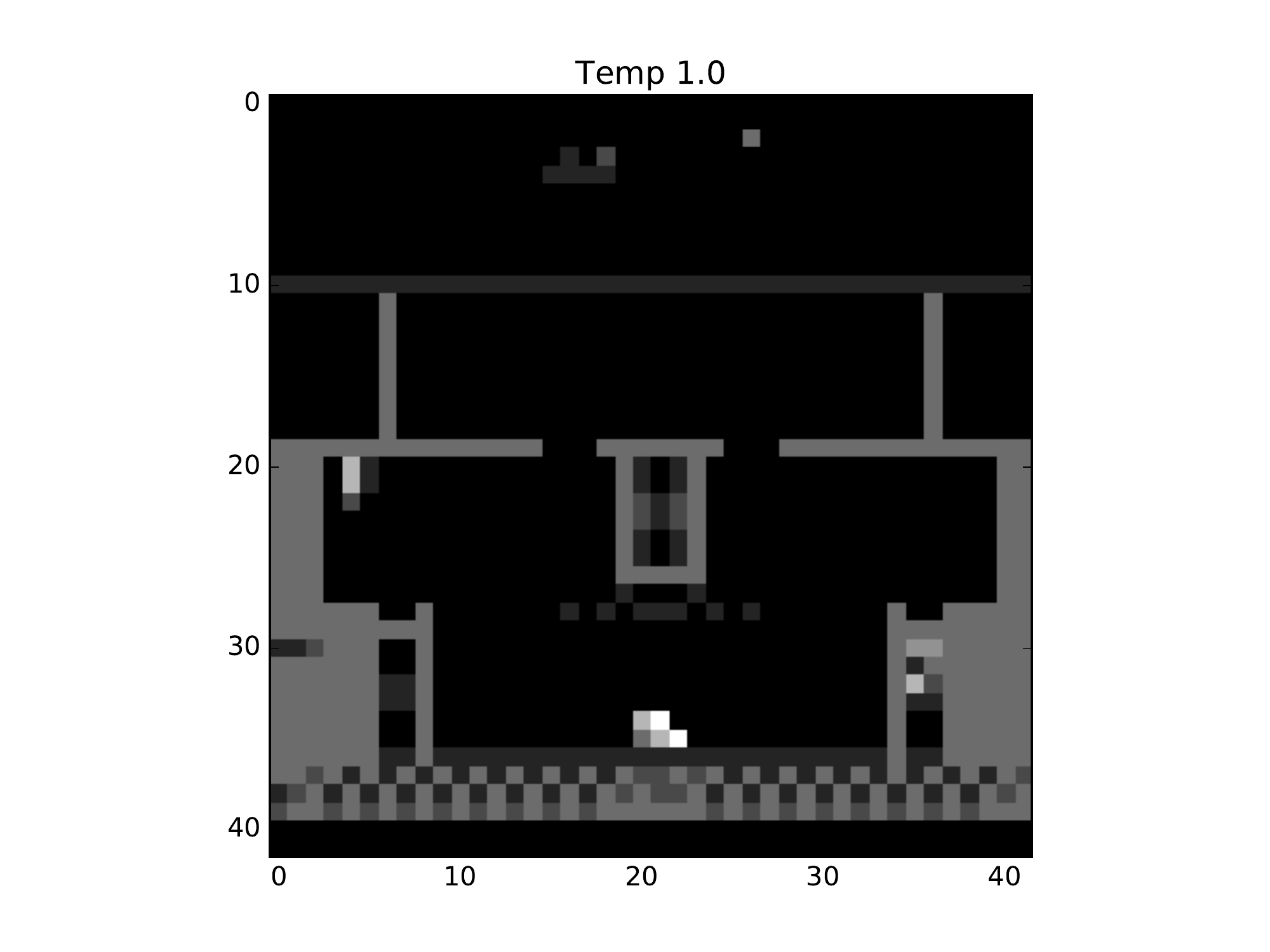}
\end{center}
\vspace{-1em}
\caption{Samples after 25K steps. \textbf{Left}: CTS, \textbf{right}: PixelCNN.}
\label{fig:samples}
\end{figure}

\subsection{Computing the Pseudo-Count}

From the previous section we obtain a particular learning rate
schedule that cannot be arbitrarily modified without deteriorating the model's
training performance or stability.
To achieve the required PG decay (b), we instead replace $\pg_n$ by $c_n \cdot \pg_n$ with a suitably decaying sequence $c_n$.

In experiments comparing actual agent performance
we empirically determined that in fact the constant learning rate $0.001$, paired with a
PG decay $c_n = c \cdot n^{-1/2}$, obtains the best exploration results
on hard exploration games like \textsc{Montezuma's Revenge}, see \figref{optim}(right).
We find the model to be robust across 1-2 orders of magnitude for the value of $c$,
and informally determine $c = 0.1$ to be a sensible configuration for achieving good
results on a broad range of Atari 2600 games (see also Section \ref{sec:pushing_exploration}).

Regarding (c), it is hard to ensure learning-positiveness for a deep neural model,
and a negative PG can occur whenever the optimizer `overshoots'
a local loss minimum. As a workaround,
we threshold the PG value at 0. To summarize, the computed pseudo-count is
\begin{equation*}
\pc_n(x) = \left( \exp\left( c \cdot n^{-1/2} \cdot (\pg_n(x))_+\right) - 1  \right) ^ {-1}.
\end{equation*}

\section{Exploration in Atari 2600 Games} \label{sec:atari}

Having described our pseudo-count friendly adaptation of PixelCNN, we now study its performance on Atari games. To this end we augment the environment reward with a pseudo-count exploration bonus, yielding the combined reward $r(x,a) + (\pc_n(x))^{-1/2}$.
As usual for neural network-based agents, we ensure the total reward lies in $[-1, 1]$ by clipping larger values.

\subsection{DQN with PixelCNN Exploration Bonus}

Our first set of experiments provides the PixelCNN exploration bonus to a DQN agent \citep{mnih2015human}\footnote{Unlike \citeauthor{bellemare16cts} we use regular Q-Learning instead of Double Q-Learning \cite{van2016deep}, as our early experiments
showed no significant advantage of
DoubleDQN with the PixelCNN-based exploration reward.}.
At each agent step, the density model receives a single frame, with which it simultaneously updates its parameters and outputs the PG.
We refer to this agent as \emph{DQN-PixelCNN}.

The \emph{DQN-CTS} agent we compare against is derived from the one in \cite{bellemare16cts}.
For better comparability,
it is trained in the same online fashion as DQN-PixelCNN,
i.e.\ the PG is computed whenever we train the
density model. By contrast, the original DQN-CTS queried
the PG at the end of each episode.

\begin{figure}[tbh]
\begin{center}
\includegraphics[width=1.60in]{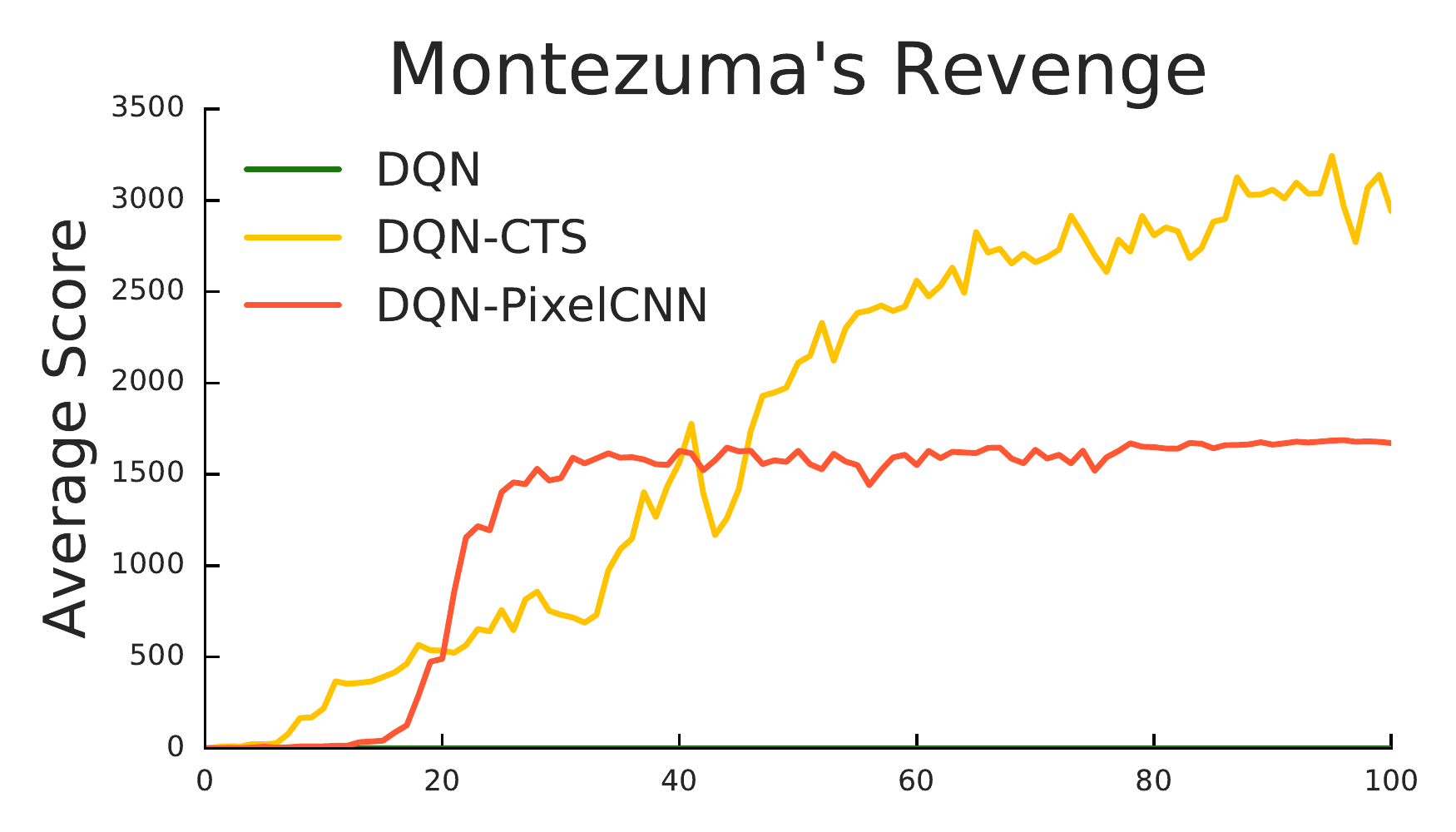}
\includegraphics[width=1.60in]{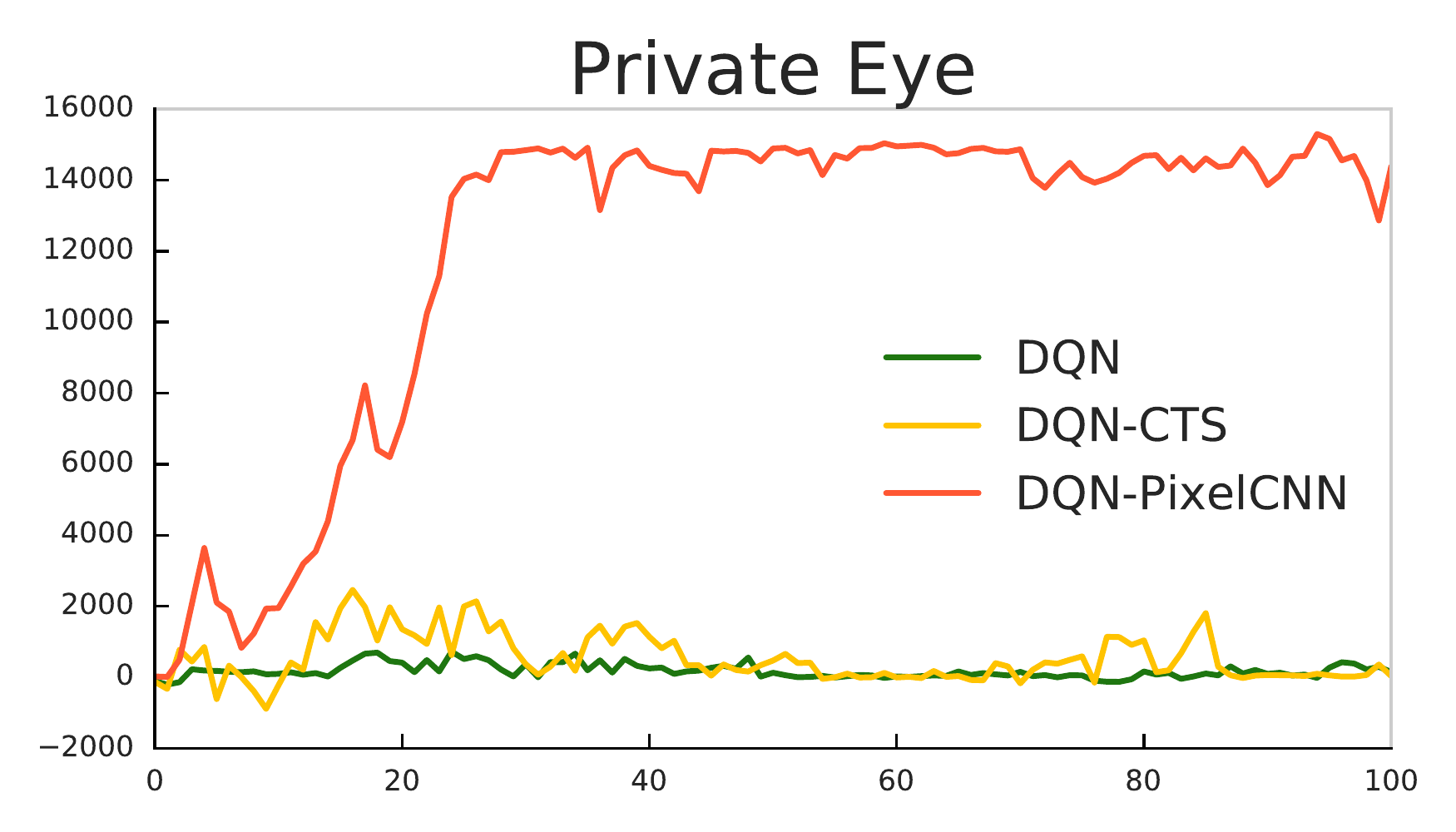}
\includegraphics[width=1.60in]{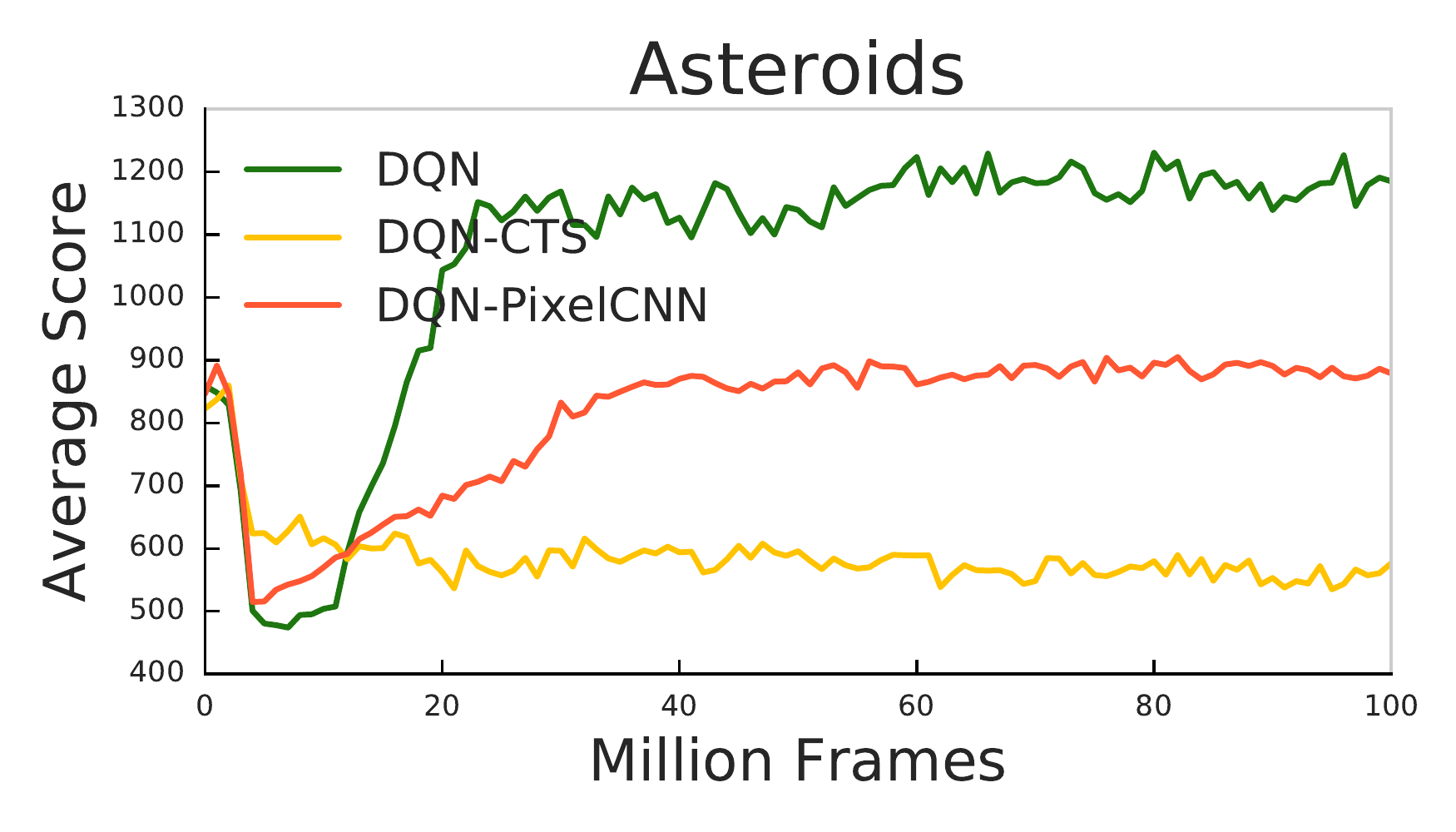}
\includegraphics[width=1.60in]{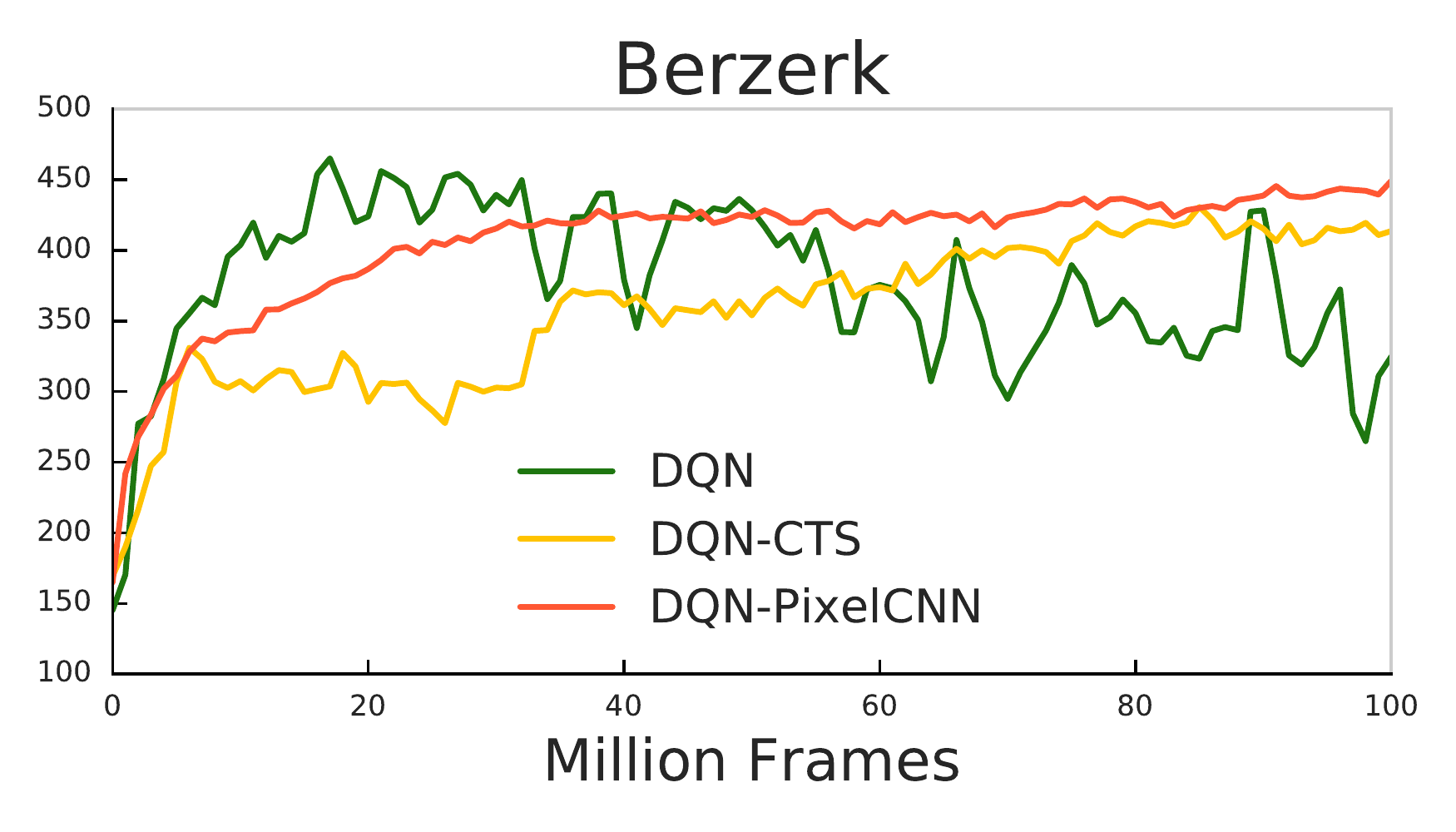}
\end{center}
\vspace{-1em}
\caption{DQN, DQN-CTS and DQN-PixelCNN on
hard exploration games (\textbf{top}) and easier ones (\textbf{bottom}).}
\label{fig:pcnn_cts}
\end{figure}

\begin{figure}[tbh]
\center{
\includegraphics[width=0.5\textwidth]{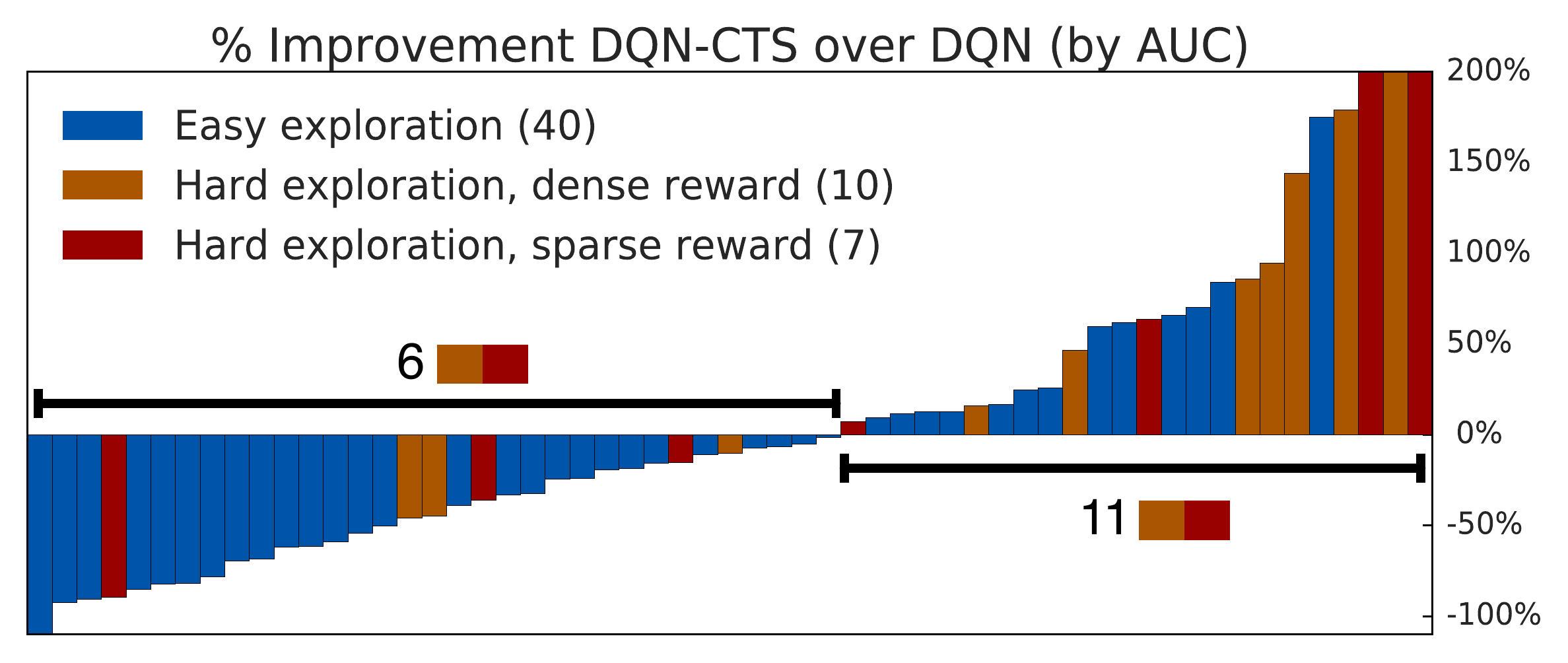}
\includegraphics[width=0.5\textwidth]{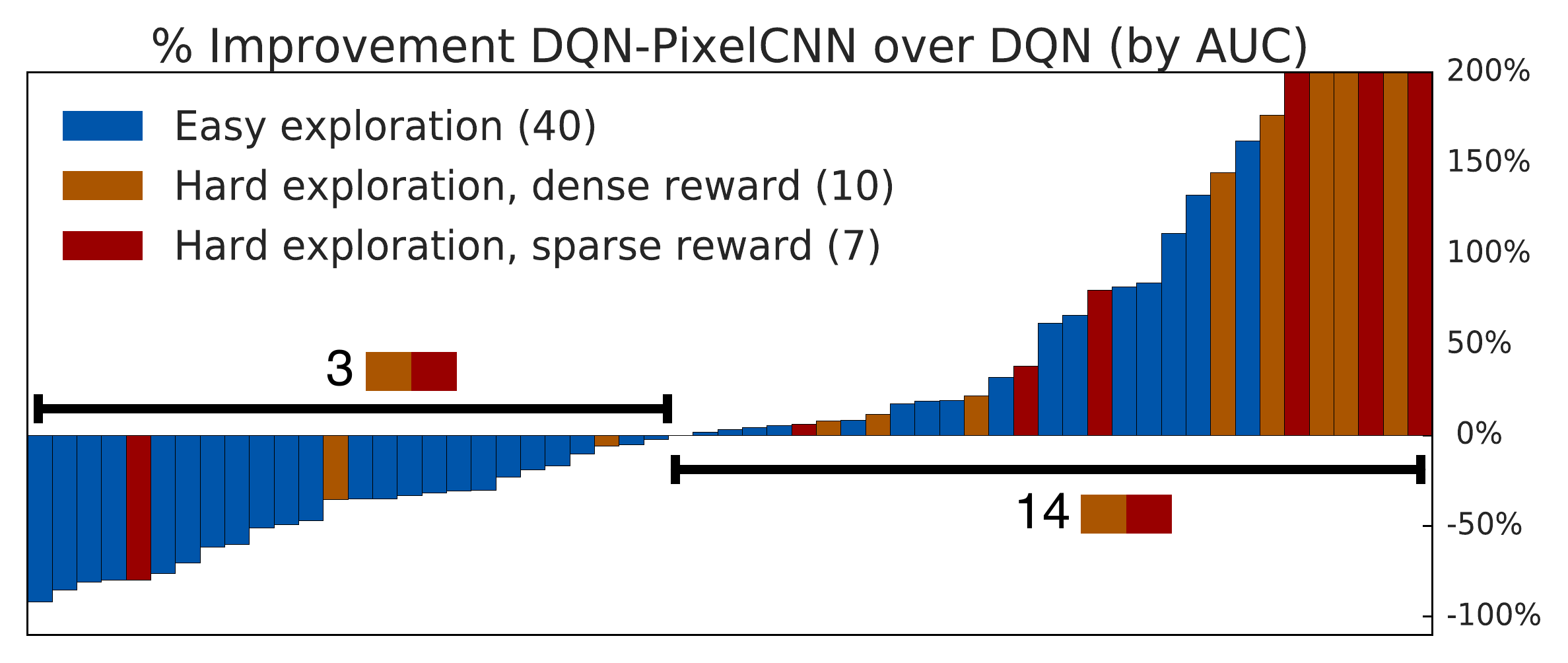}
}
\vspace{-2em}
\caption{Improvements (in \% of AUC) of DQN-PixelCNN and DQN-CTS over DQN in 57 Atari games. Annotations indicate the number of hard exploration games with positive (right) and negative (left) improvement, respectively.}
\label{fig:pcnn_cts_improv}
\end{figure}

Unless stated otherwise, we always use the mixed Monte Carlo update (MMC)
for the intrinsically motivated agents\footnote{The use of MMC in a replay-based agent poses a minor complication, as the MC return is not available for replay until the end of an episode. For simplicity, in our implementation we disregard this detail and set the MC return to 0 for transitions from the most recent episode.}, but regular Q-Learning for the baseline DQN.

\figref{pcnn_cts} shows training curves of DQN compared
to DQN-CTS and DQN-PixelCNN.
On the famous \textsc{Montezuma's Revenge}, both intrinsically motivated
agents vastly outperform the baseline DQN. On other hard exploration games
(\textsc{Private Eye}; or \textsc{Venture}, appendix \figref{pcnn_cts_full}),
DQN-PixelCNN achieves state of the art results, substantially outperforming DQN and DQN-CTS.
The other two games shown (\textsc{Asteroids}, \textsc{Berzerk}) pose easier
exploration problems, where the reward bonus
should not provide large improvements and may have
a negative effect by skewing the reward landscape.
Here, DQN-PixelCNN behaves more gracefully and still outperforms DQN-CTS. We hypothesize this is due to a qualitative difference between the models, see Section \ref{sec:quality}.

Overall PixelCNN provides the DQN agent with a larger advantage than CTS,
and often accelerates or stabilizes training
even when not affecting peak performance. Out of 57 Atari games,
DQN-PixelCNN outperforms DQN-CTS in 52 games by maximum achieved score, and
51 by AUC (methodology in Appendix \ref{sec:methodology}). See \figref{pcnn_cts_improv} for a high
level comparison (appendix \figref{pcnn_cts_full}
for full training graphs).
The greatest gains from using either exploration bonus
are observed in games categorized as \textit{hard exploration} games in the
`taxonomy of exploration' in \citep[][reproduced in Appendix \ref{sec:taxonomy}]{bellemare16cts}, specifically
in the most challenging \textit{sparse reward} games (e.g.\
\textsc{Montezuma's Revenge}, \textsc{Private Eye}, \textsc{Venture}).

\subsection{A Multi-Step RL Agent with PixelCNN}

Empirical practitioners know that techniques beneficial for one
agent architecture often can be detrimental for a different algorithm.
To demonstrate the wide applicability of the PixelCNN
exploration bonus, we also evaluate it with the
more recent \textit{Reactor} agent\footnote{The
exact agent variant is referred to as `$\beta$-LOO' with $\beta = 1$.} \cite{gruslys17reactor}.
This replay-based actor-critic agent represents its policy and value function by a recurrent neural network
and, crucially, uses the multi-step Retrace($\lambda$) algorithm
for policy evaluation, replacing the MMC we use in DQN-PixelCNN.

To reduce impact on computational
efficiency of this agent, we sub-sample intrinsic rewards:
we perform updates of the PixelCNN model and compute the reward
bonus on (randomly chosen) 25\% of all steps, leaving the agent's reward
unchanged on other steps. We use the same PG decay schedule of
$0.1 n^{-1/2}$, with $n$ the number of model updates.

\begin{figure}[tbh]
\center{
\includegraphics[width=1.6in]{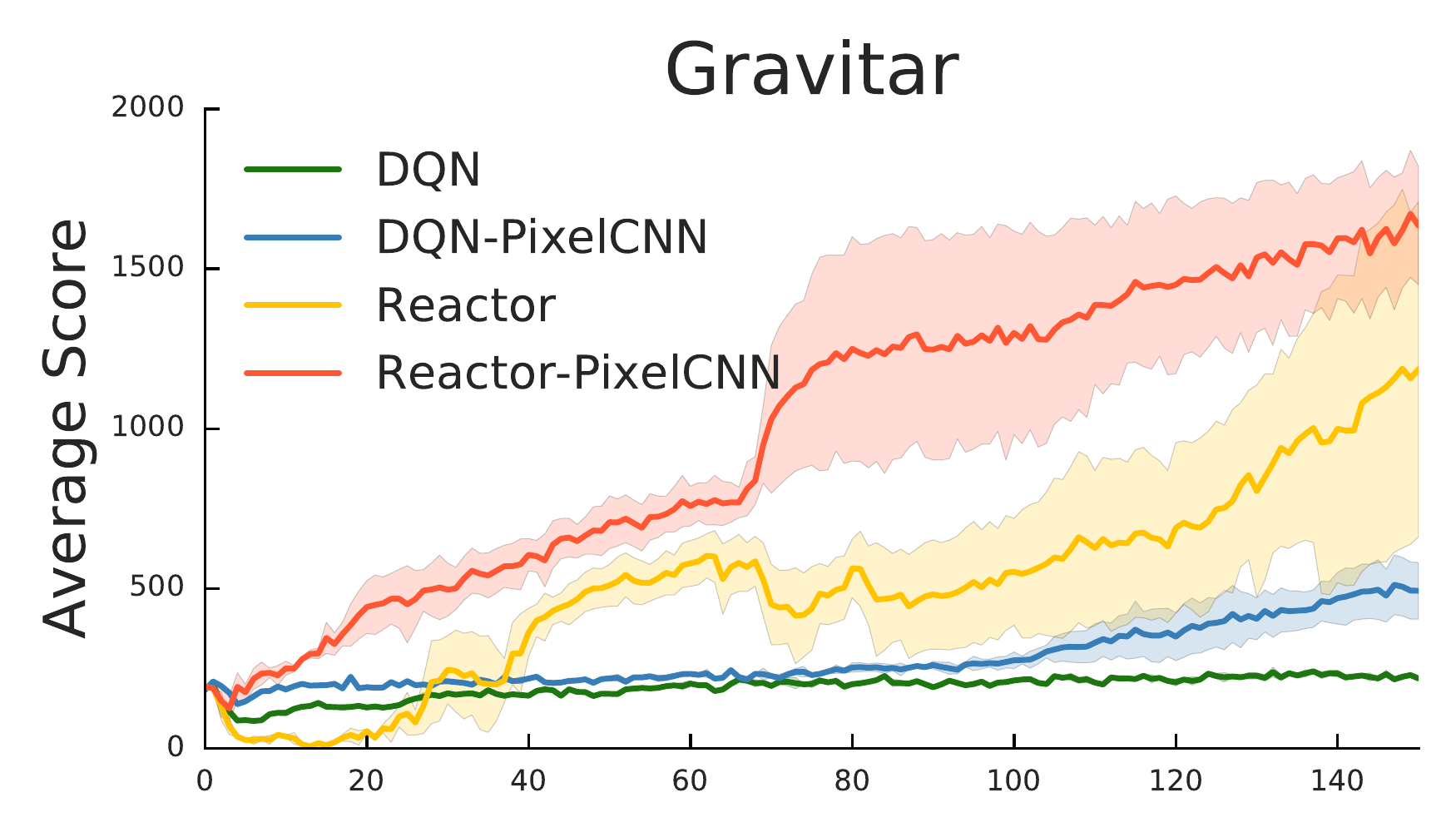}
\includegraphics[width=1.6in]{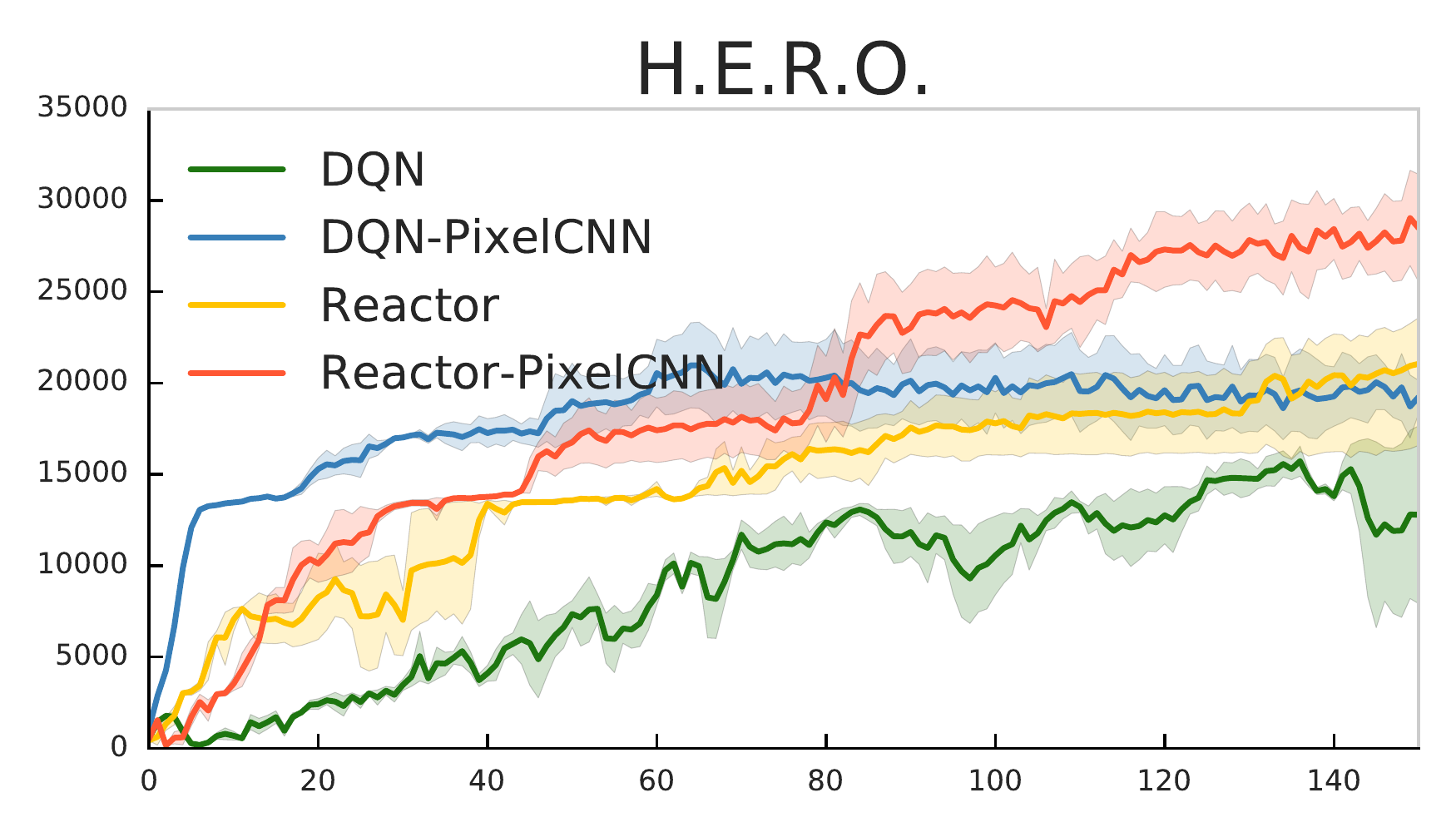}
\includegraphics[width=1.6in]{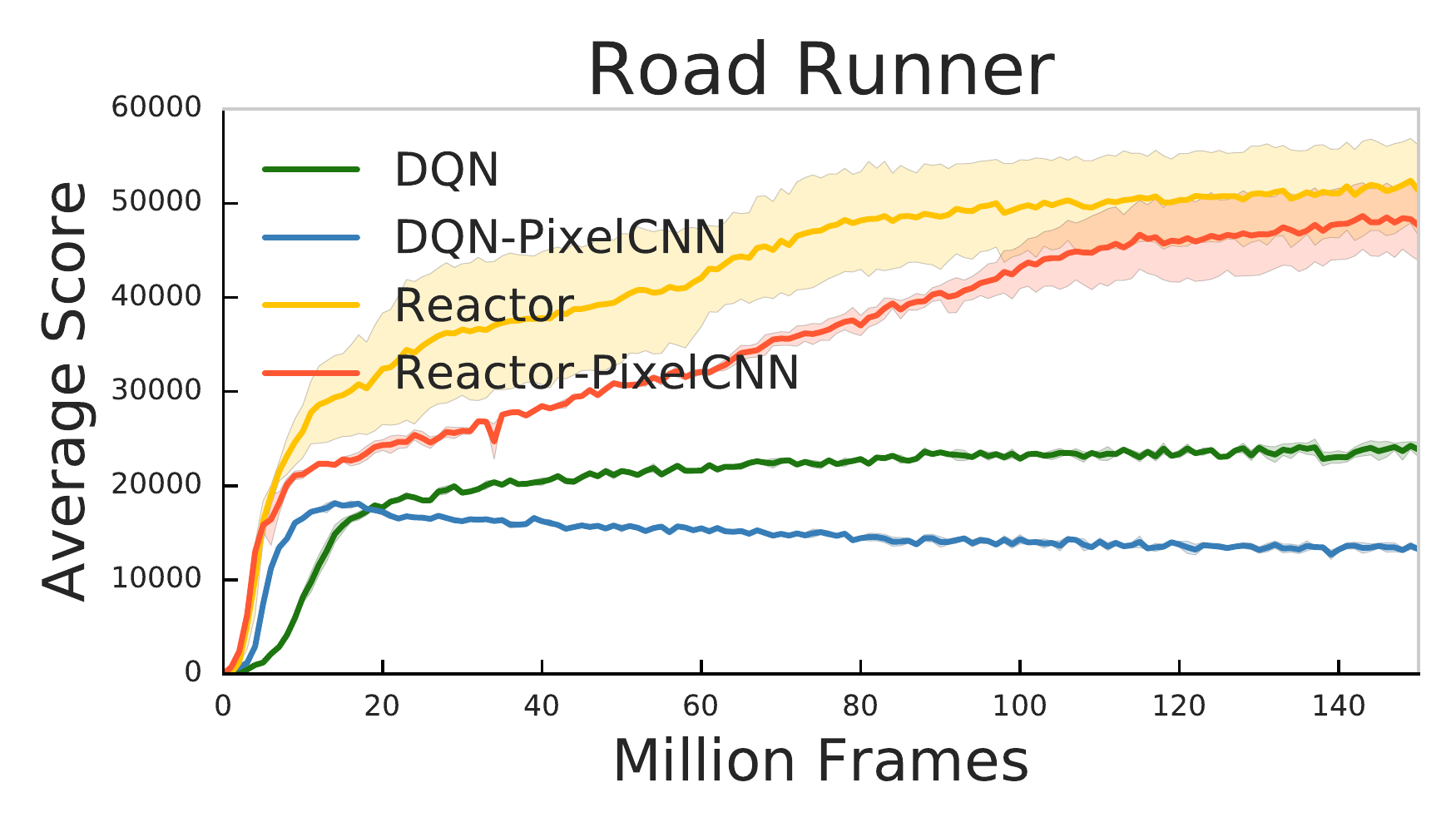}
\includegraphics[width=1.6in]{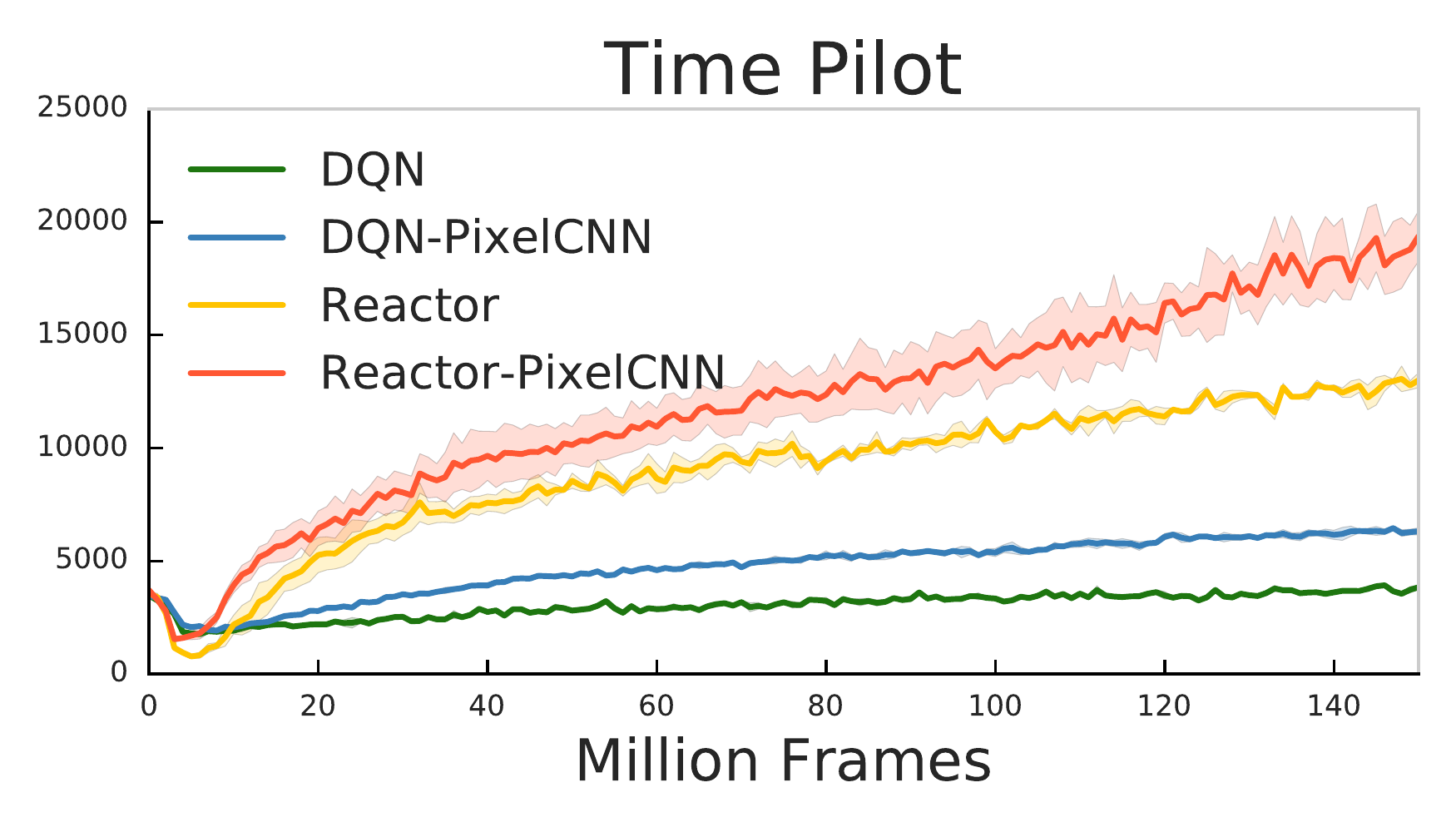}
}
\vspace{-2em}
\caption{Reactor/Reactor-PixelCNN and DQN/DQN-PixelCNN training performance (averaged over 3 seeds).
\label{fig:reactor}}
\end{figure}

Training curves for the Reactor/Reactor-PixelCNN agent compared
to DQN/DQN-PixelCNN are shown in \figref{reactor}.
The baseline Reactor agent is superior to the DQN agent, obtaining
higher scores and learning faster in about 50 out of 57 games.
It is further improved on a large fraction of games by the
PixelCNN exploration reward, see \figref{reactor_improv}
(full training graphs in appendix \figref{reactor_full}).

The effect of the exploration bonus
is rather uniform, yielding improvements on a broad range of games.
In particular, Reactor-PixelCNN enjoys better sample efficiency
(in terms of area under the curve, AUC) than vanilla Reactor.
We hypothesize that, like other policy gradient algorithms, Reactor
generally suffers from weaker exploration than its value-based
counterpart DQN. This aspect is much helped by the exploration bonus, boosting the agent's sample efficiency in many environments.

\begin{figure}[tbh]
\center{
\includegraphics[width=0.5\textwidth]{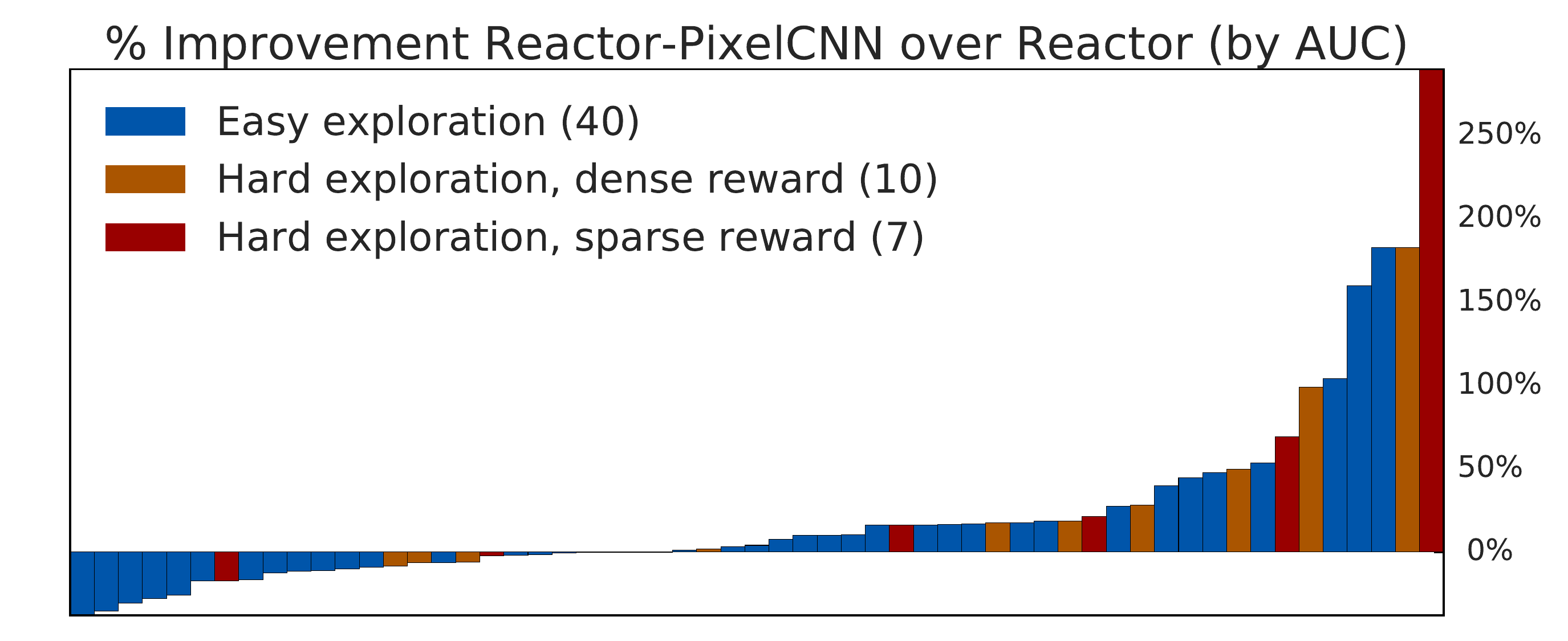}
}
\vspace{-2em}
\caption{Improvements (in \% of AUC) of Reactor-PixelCNN over Reactor in 57 Atari games.}
\label{fig:reactor_improv}
\end{figure}

However, on hard exploration games with sparse rewards,
Reactor seems unable to make full use of the exploration bonus.
We believe this is because, in very sparse settings, the propagation of reward information across long horizons becomes crucial. The MMC takes one extreme of this view, directly learning from the observed returns. The Retrace$(\lambda)$ algorithm, on the other hand, has an effective horizon which depends on $\lambda$ and, critically, the truncated importance sampling ratio. This ratio results in the discarding of trajectories which are off-policy, i.e.\ unlikely under the current policy. We hypothesize that the very goal of the Retrace($\lambda$) algorithm to learn cautiously is what prevents it from taking full advantage of the exploration bonus!

\section{Quality of the Density Model}\label{sec:quality}

PixelCNN can be expected to be more expressive and accurate
than the less advanced CTS model, and indeed, samples generated
after training are somewhat higher quality (\figref{samples}).
However, we
are not using the generative function of the models when
computing an exploration bonus, and a better generative model
does not necessarily give rise to better probability estimates
\cite{theis2015note}.

\begin{figure}[tbh]
\center{
\includegraphics[width=1.5in]{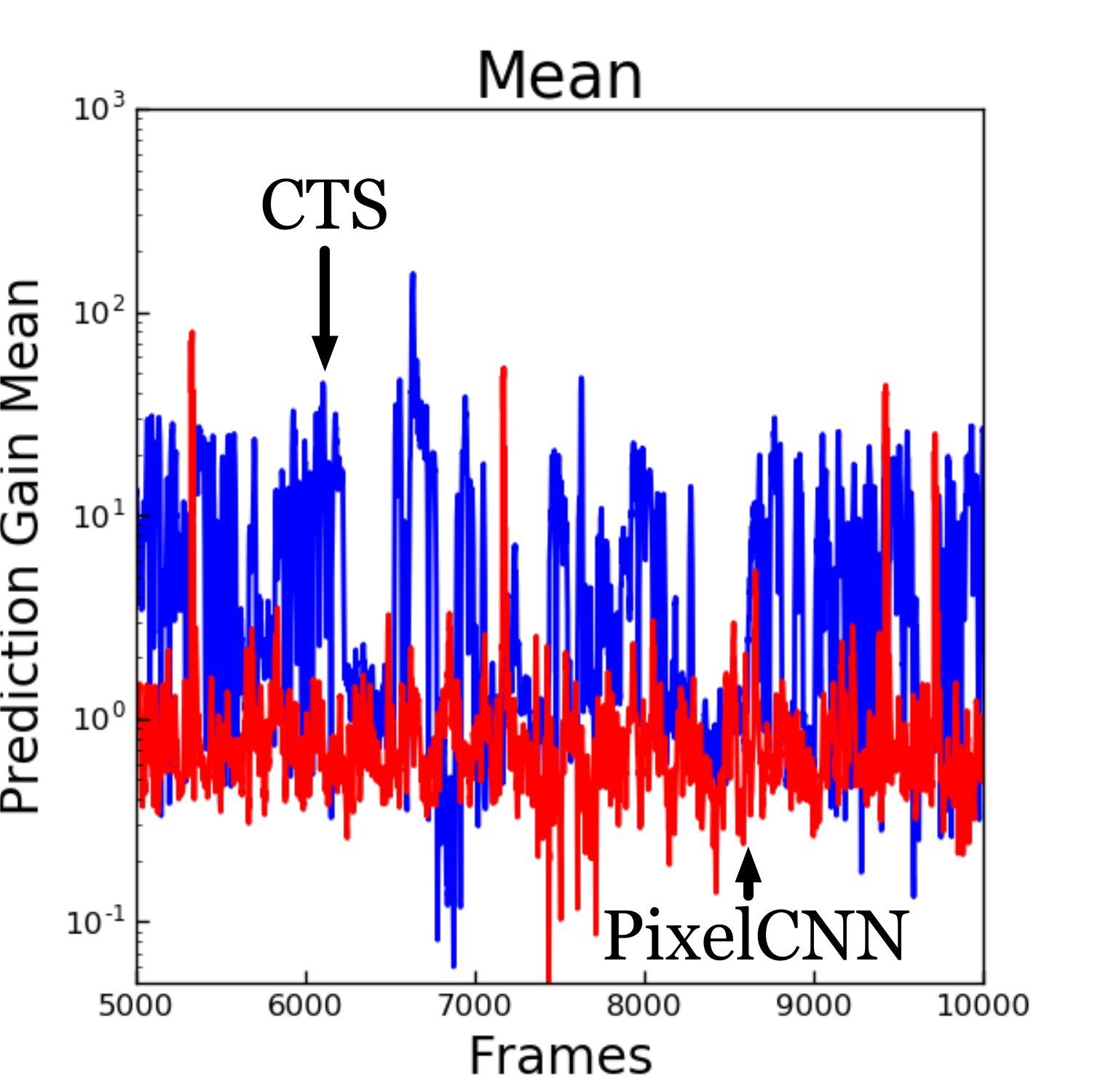}
\includegraphics[width=1.5in]{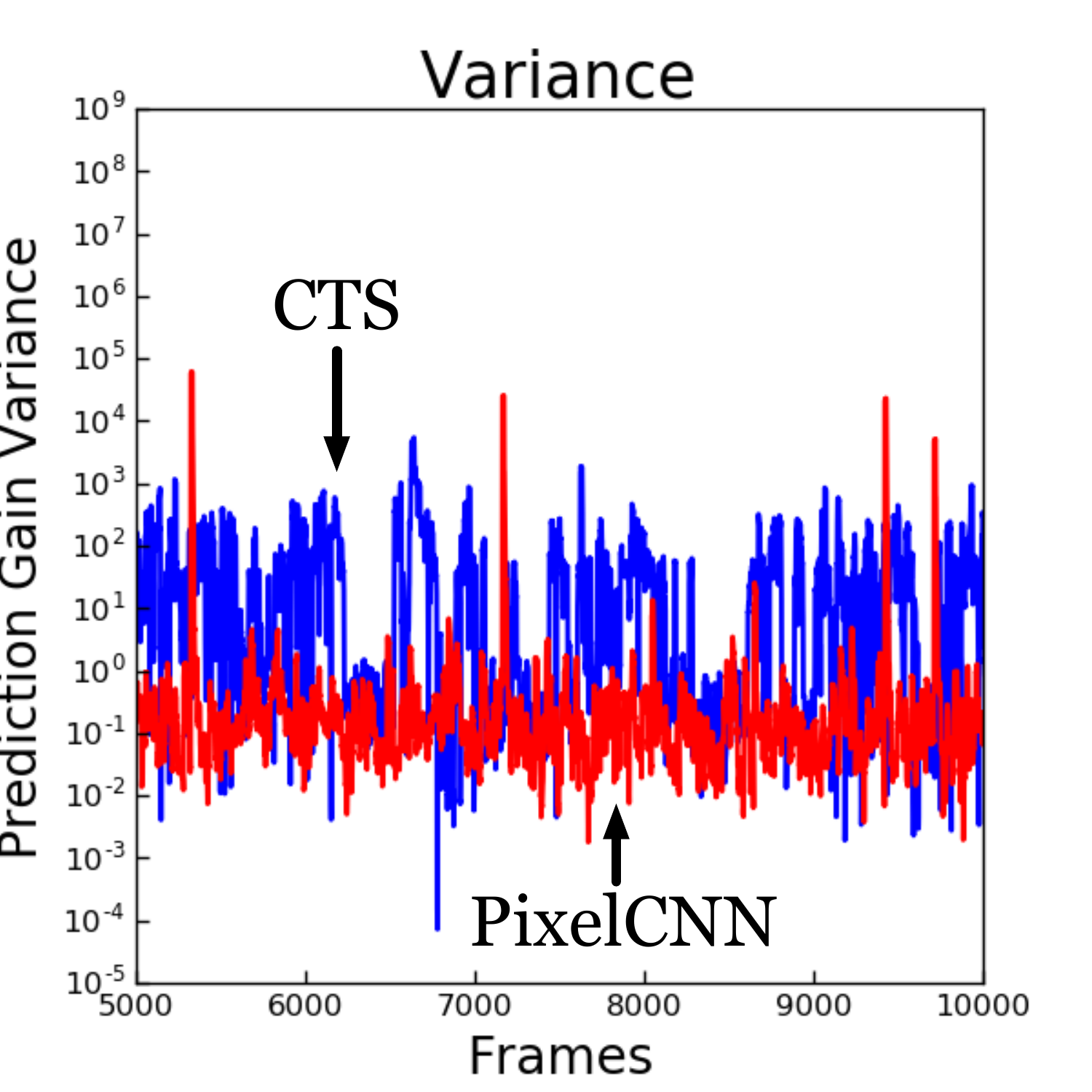}
\vspace{-1.em}
}
\caption{PG on \textsc{Montezuma's Revenge} (log scale).}
\label{fig:cts_pcnn_pg}
\end{figure}

In \figref{cts_pcnn_pg} we compare the PG produced by
the two models throughout 5K training steps. PixelCNN
consistently produces PGs lower than CTS. More importantly, its
PGs are \emph{smoother}, exhibiting less variance between successive states,
while showing more pronounced peaks at certain infrequent events. This yields
a reward bonus that is less harmful in easy exploration games,
while
providing a strong signal in the case of novel or rare events.

\begin{figure}[htb]
\center{
\includegraphics[width=2.6in]{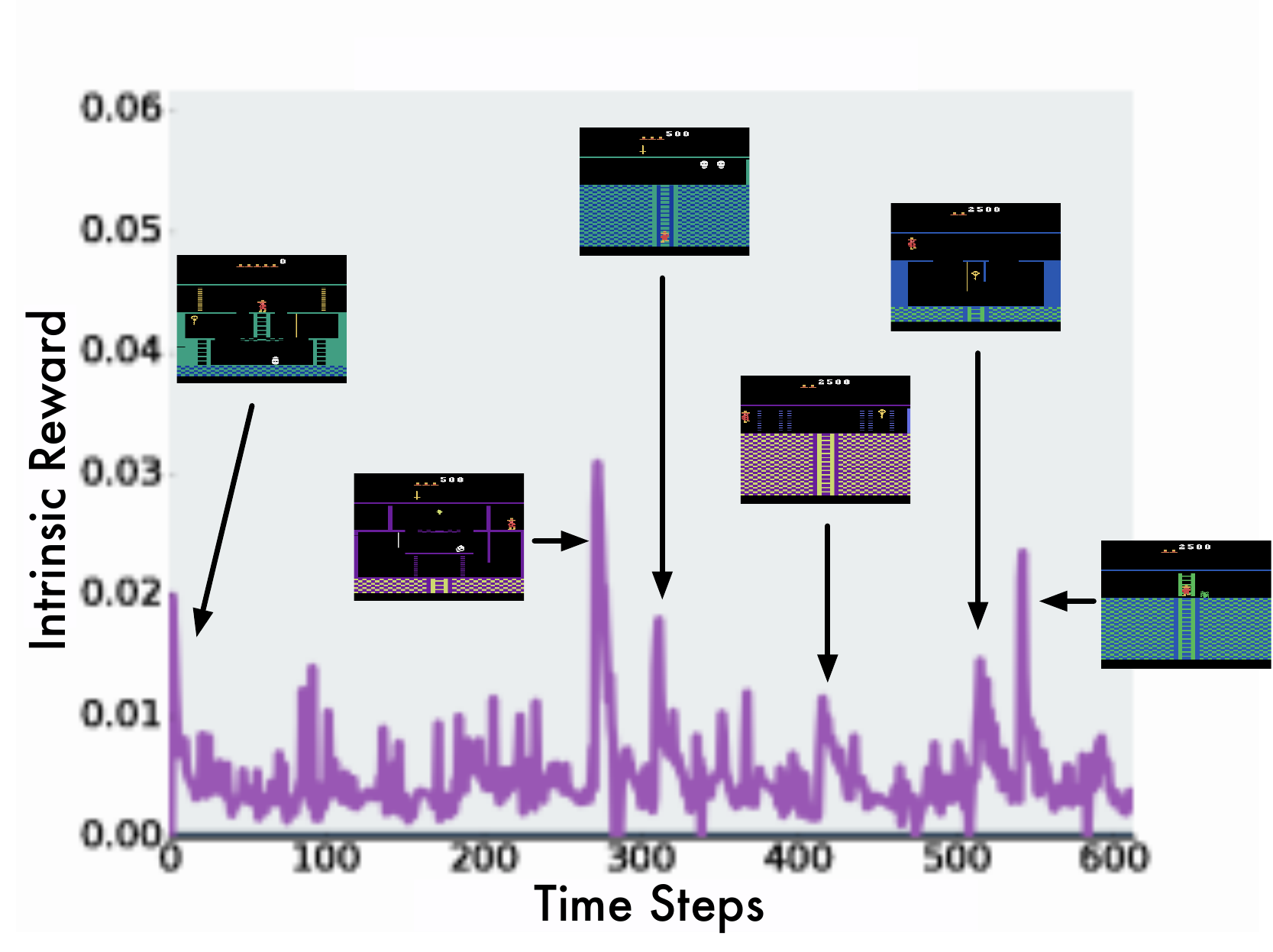}
}
\vspace{-1em}
\caption{Intrinsic reward in \textsc{Montezuma's Revenge}.\label{fig:montezuma_revenge_prediction_gain}}
\end{figure}

Another distinguishing feature of PixelCNN is its non-decaying step-size.
The per-step PG never completely vanishes, as the model tracks the most
recent data. This provides
an unexpected benefit: the agent remains mildly
surprised by significant state changes,
e.g.\ switching rooms in \textsc{Montezuma's Revenge}.
These persistent rewards act as \emph{milestones}
that the agent learns to return to. This is illustrated in
\figref{montezuma_revenge_prediction_gain},
depicting the intrinsic reward over the course of an episode.
The agent routinely revisits the
right-hand side of the torch room, not because it leads to
reward but just to ``take in the sights''.
A video of the episode is provided at
\url{http://youtu.be/232tOUPKPoQ}.\footnote{Another agent video on the game \textsc{Private Eye} can be found at \url{http://youtu.be/kNyFygeUa2E}.}

Lastly, PixelCNN's convolutional nature is expected
to be beneficial for its sample efficiency.
In Appendix \ref{sec:convcts} we compare to
a convolutional CTS and confirm that this explains part, but
not all of PixelCNN's advantage over vanilla CTS.

\section{Importance of the Monte Carlo Return}

\begin{figure}[tbh]
\center{
\includegraphics[width=1.6in]{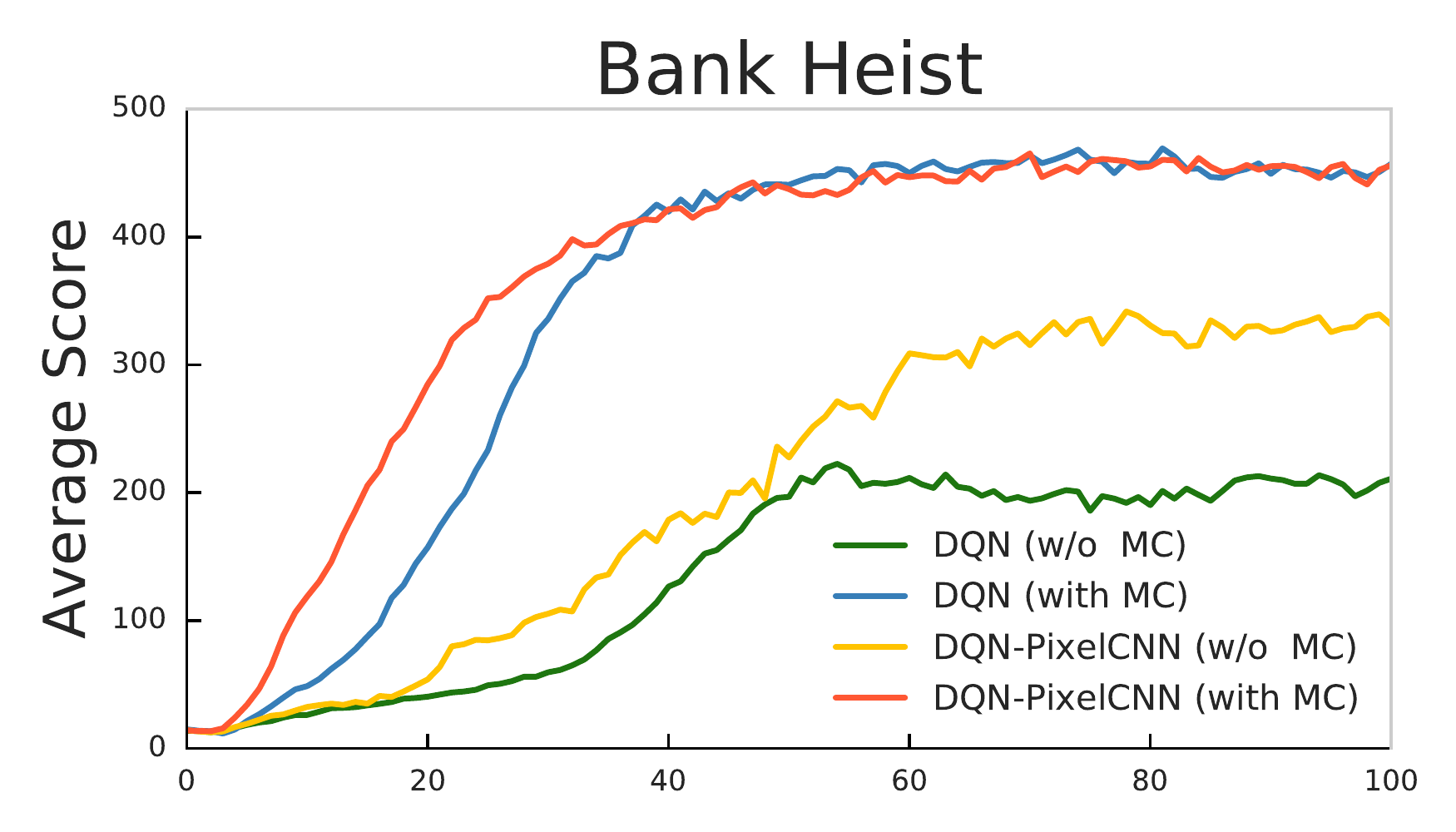}
\includegraphics[width=1.6in]{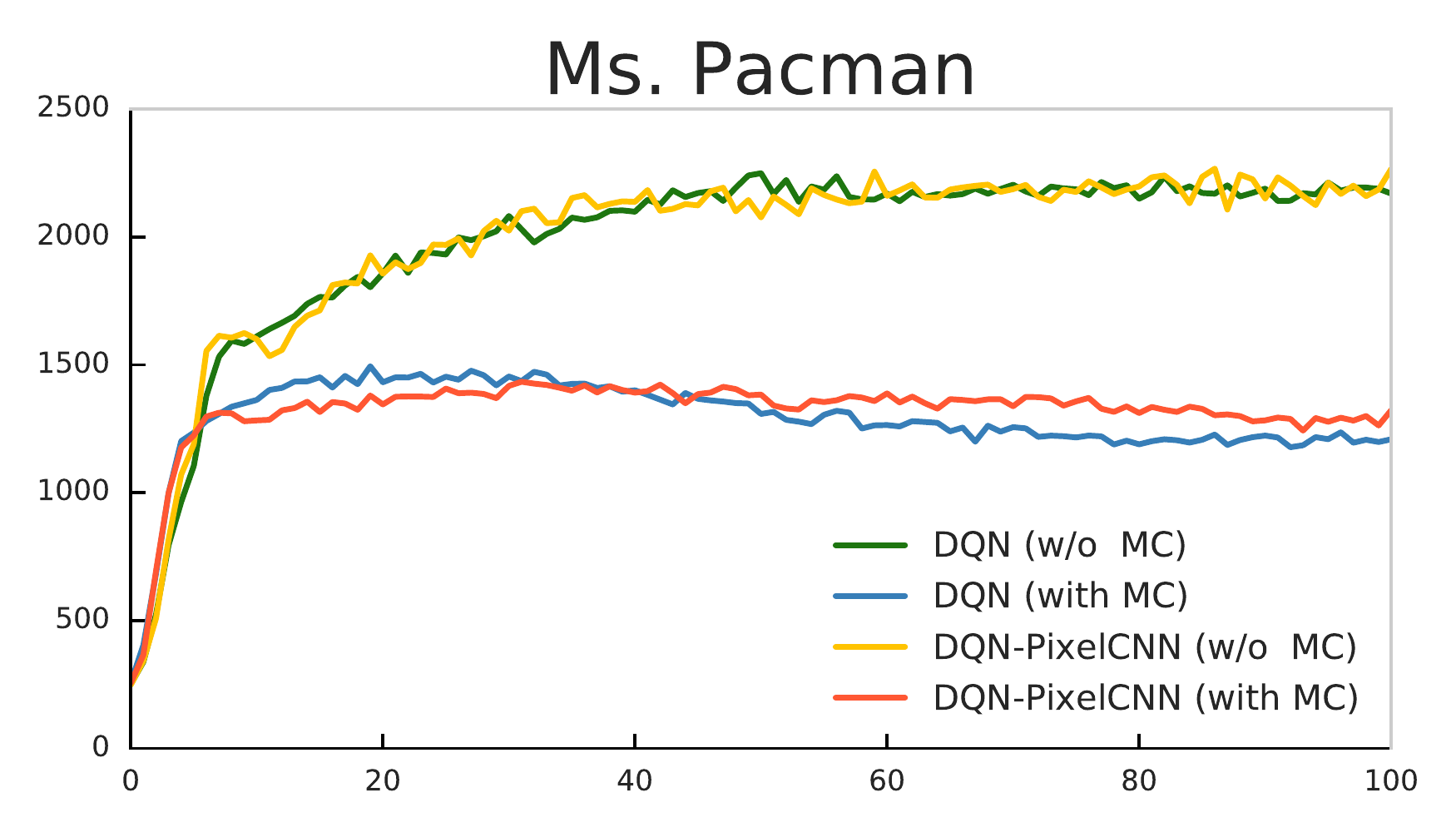}
\includegraphics[width=1.6in]{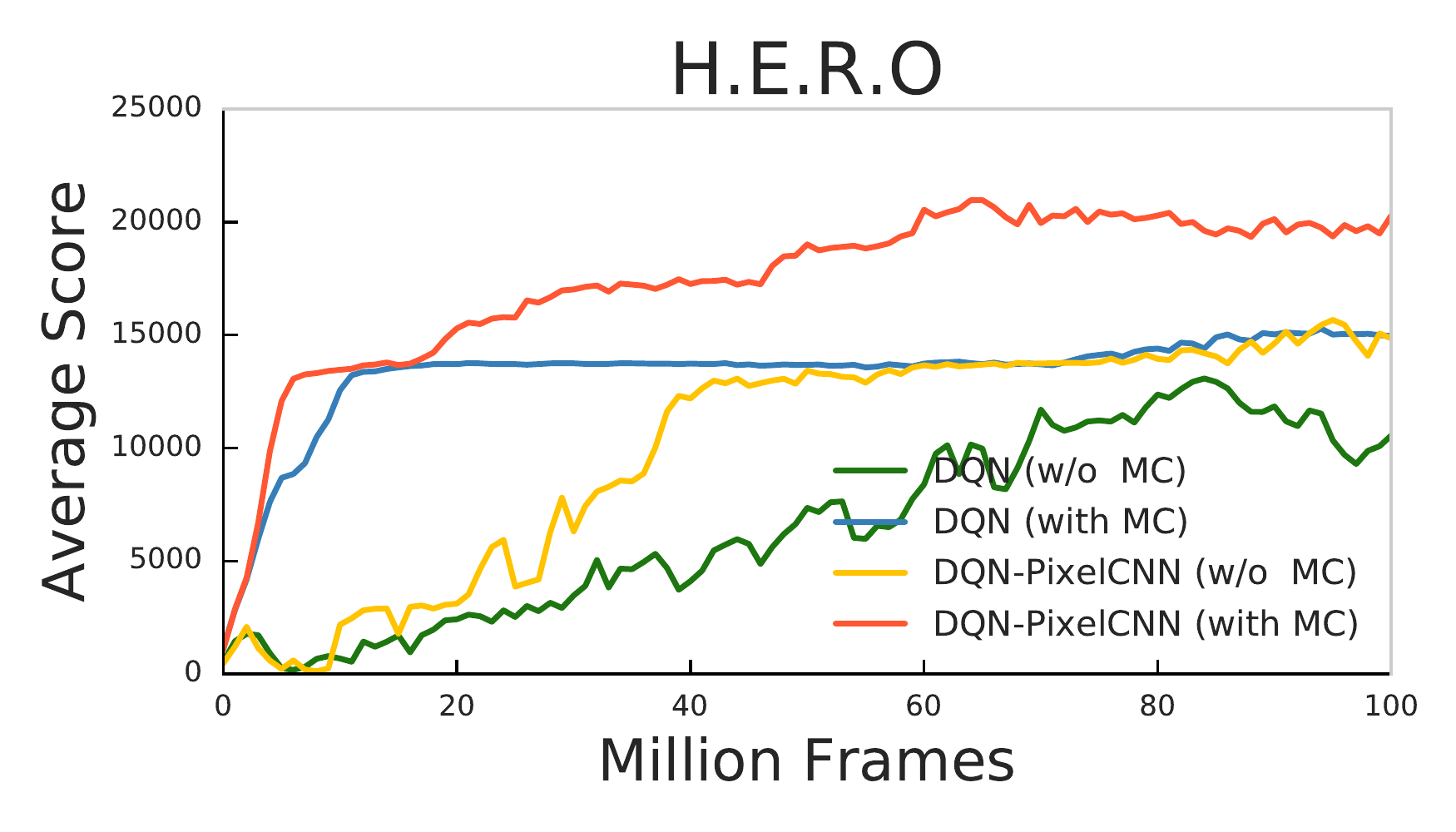}
\includegraphics[width=1.6in]{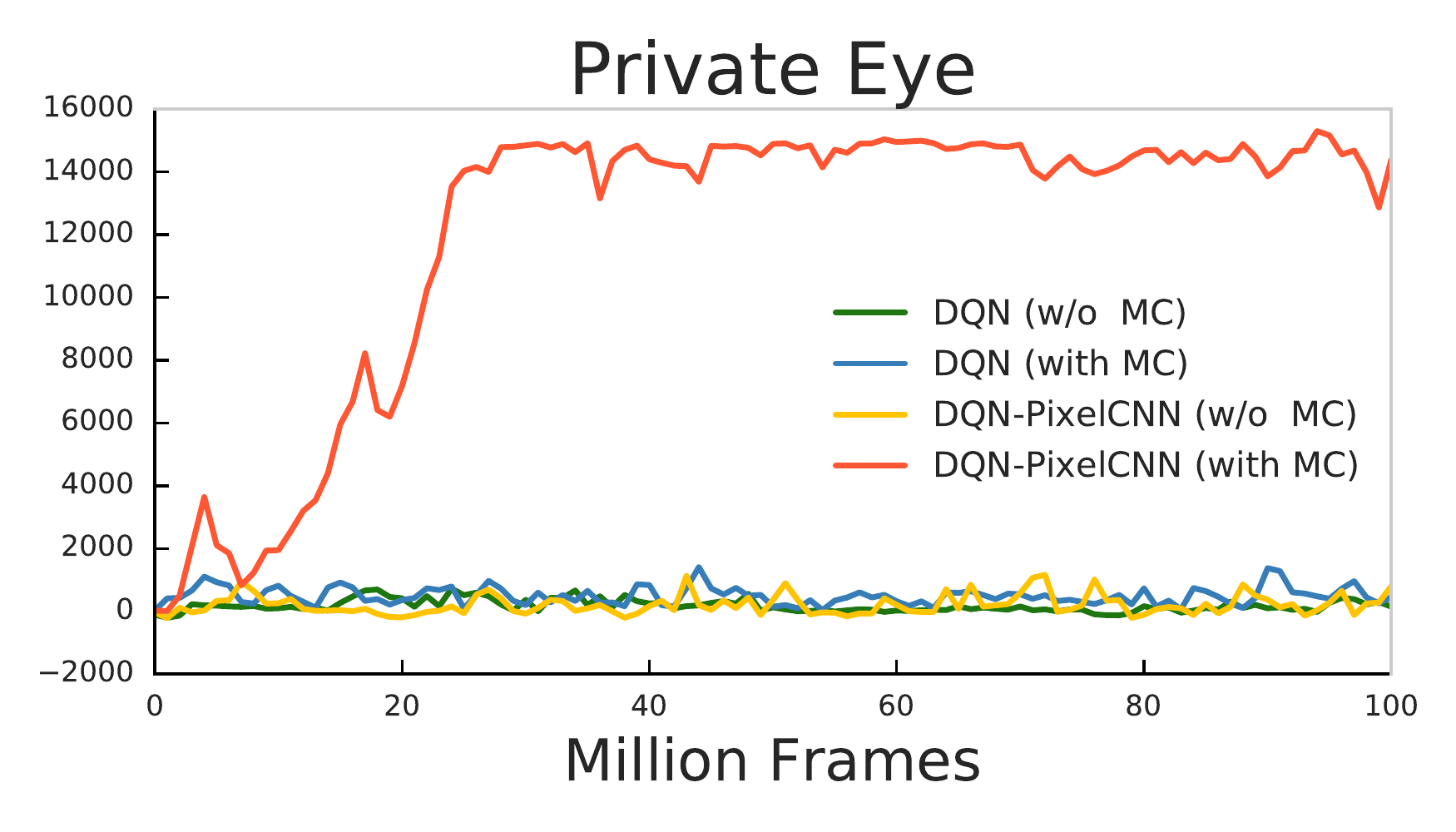}
}
\vspace{-2em}
\caption{\textbf{Top}: games where MMC completely explains the
improved/decreased performance of DQN-PixelCNN compared to DQN. \textbf{Bottom-left}: MMC and PixelCNN show additive benefits. \textbf{Bottom-right}: hard exploration, sparse
reward game -- only combining MMC and PixelCNN bonus achieves
training progress.
\label{fig:mc_games}}
\end{figure}

Like for DQN-CTS, the success of DQN-PixelCNN hinges on
the use of the mixed Monte Carlo update.
The transient and vanishing nature of the exploration rewards requires the learning
algorithm to latch on to these rapidly. The MMC
serves this end as a simple multi-step method, helping to
propagate reward information faster.
An additional benefit
lies in the fact that the Monte Carlo return helps bridging long horizons in
environments where rewards are far apart and encountered rarely.
On the other hand,
it is important to note that the Monte Carlo return's on-policy nature
increases variance in the learning algorithm, and can prevent the algorithm's
convergence to the optimal policy when training off-policy.
It can therefore be expected to adversely affect training performance in
some games.

To distill the effect of the MMC on performance,
we compare all four combinations of DQN with/without PixelCNN exploration bonus
and with/without MMC. \figref{mc_games} shows the performance
of these four agent variants (graphs for all games
are shown in \figref{mc_full}).
These games were picked to illustrate several commonly occurring cases:
\begin{itemize}
 \item MMC speeds up training and improves
 final performance significantly (examples: \textsc{Bank Heist}, \textsc{Time Pilot}).
 In these games, MMC alone explains most or all of the improvement
 of DQN-PixelCNN over DQN.

 \item MMC hurts performance
 (examples: \textsc{Ms.\ Pac-Man}, \textsc{Breakout}).
 Here too, MMC alone explains most of the difference
 between DQN-PixelCNN and DQN.

 \item MMC and PixelCNN reward bonus
 have a compounding effect (example: \textsc{H.E.R.O.}).

\end{itemize}

\begin{figure*}[tbh]
\center{
\includegraphics[width=2.22in]{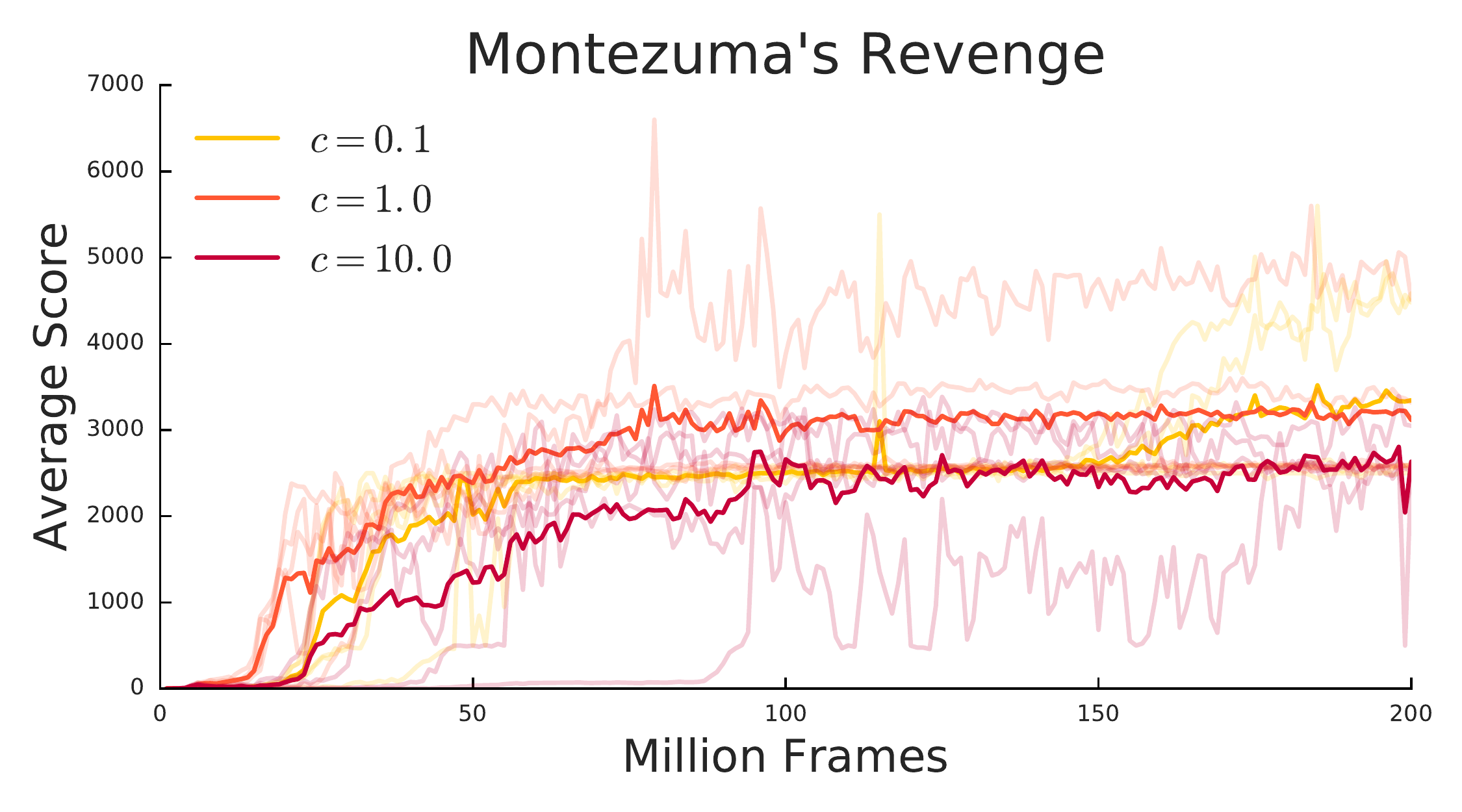}
\includegraphics[width=2.22in]{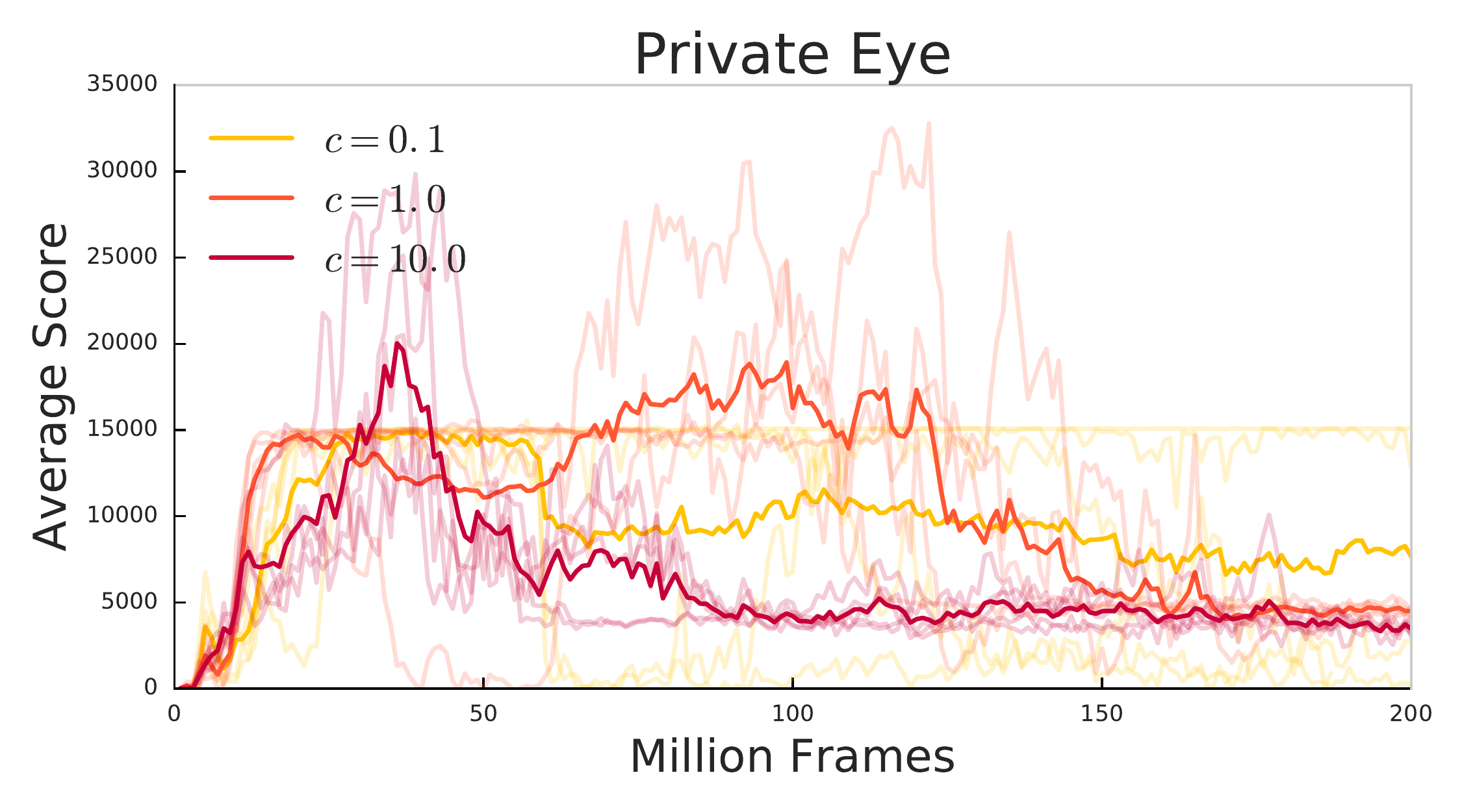}
\includegraphics[width=2.22in]{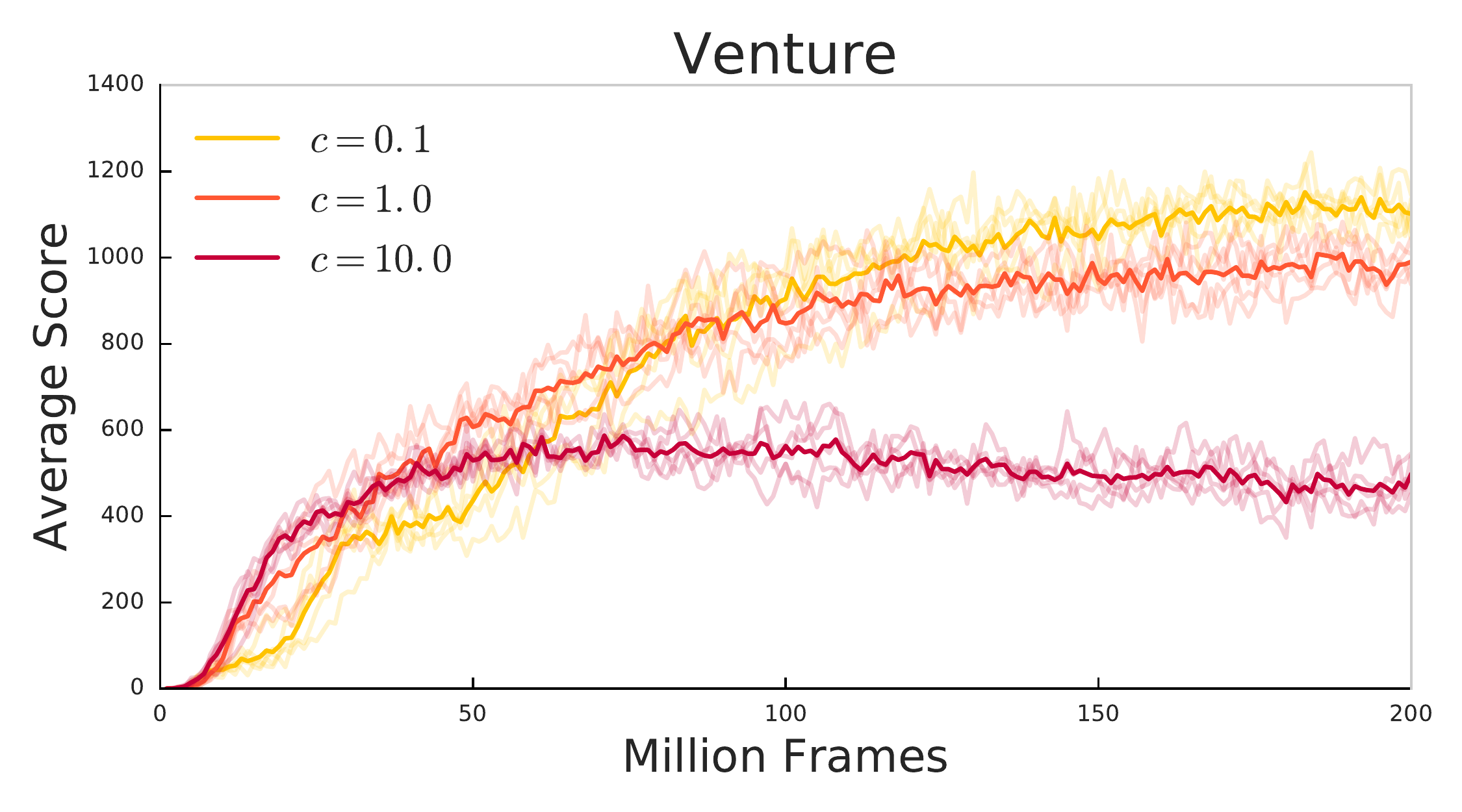}
}
\vspace{-2em}
\caption{DQN-PixelCNN, hard exploration games, different PG scales
$c \cdot n^{-1/2}\cdot \pg_n$ ($c = 0.1, 1, 10$)
(5 seeds each).}
\label{fig:pcnn_scales}
\end{figure*}

Most importantly, the situation is rather different when we restrict our
attention to the hardest exploration games with sparse rewards.
Here the baseline DQN agent
fails to make any training progress, and neither Monte Carlo return
nor the exploration bonus alone provide any significant benefit. Their combination
however grants the agent rapid training progress and allows it to achieve high
performance.

One effect of the exploration bonus in these games
is to provide a denser reward landscape,
enabling the agent to learn meaningful policies.
Due to the transient nature of the exploration bonus,
the agent needs to be able to learn from this reward signal faster
than regular one-step methods allow, and MMC proves to be
an effective solution.

\section{Pushing the Limits of Intrinsic Motivation} \label{sec:pushing_exploration}

In this section we explore the idea of a `maximally curious' agent,
whose reward function is dominated by the exploration bonus.
For that we increase the PG scale, previously chosen conservatively to
avoid adverse effects on easy exploration games.

\figref{pcnn_scales} shows DQN-PixelCNN performance on the hardest exploration
games when the PG scale is increased by 1-2 orders of magnitude.
The algorithm seems fairly robust across a wide
range of scales: the main effect of increasing this parameter is to trade off
exploration (seeking maximal reward) with exploitation (optimizing the current policy).

As expected, a higher PG scale translates to
stronger exploration: several runs
obtain record peak scores (900 in \textsc{Gravitar}, 6,600 in \textsc{Montezuma's Revenge},
39,000 in \textsc{Private Eye}, 1,500 in \textsc{Venture}) surpassing
the state of the art by a substantial margin (for previously published results, see Appendix \ref{sec:taxonomy}).
Aggressive scaling
speeds up the agent's exploration and achieves peak performance rapidly,
but can also deteriorate its stability and long-term performance.
Note that in practice, because of the non-decaying step-size the PG
does not vanish. After reward clipping, an overly inflated exploration bonus can
therefore become essentially constant, no longer providing a useful
intrinsic motivation signal to the agent.

\begin{figure}[tbh]
\center{
\includegraphics[width=2.8in]{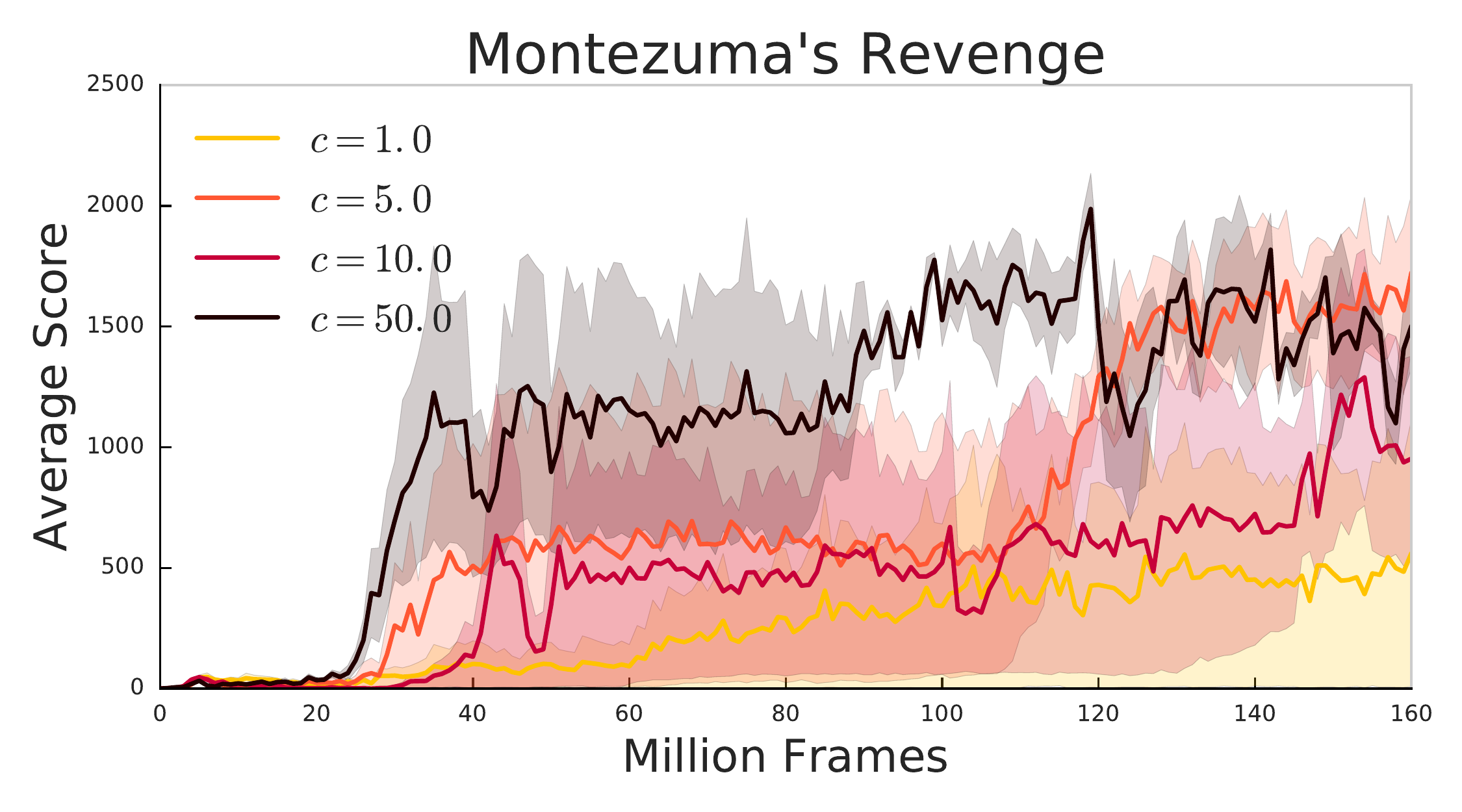}
\includegraphics[width=2.8in]{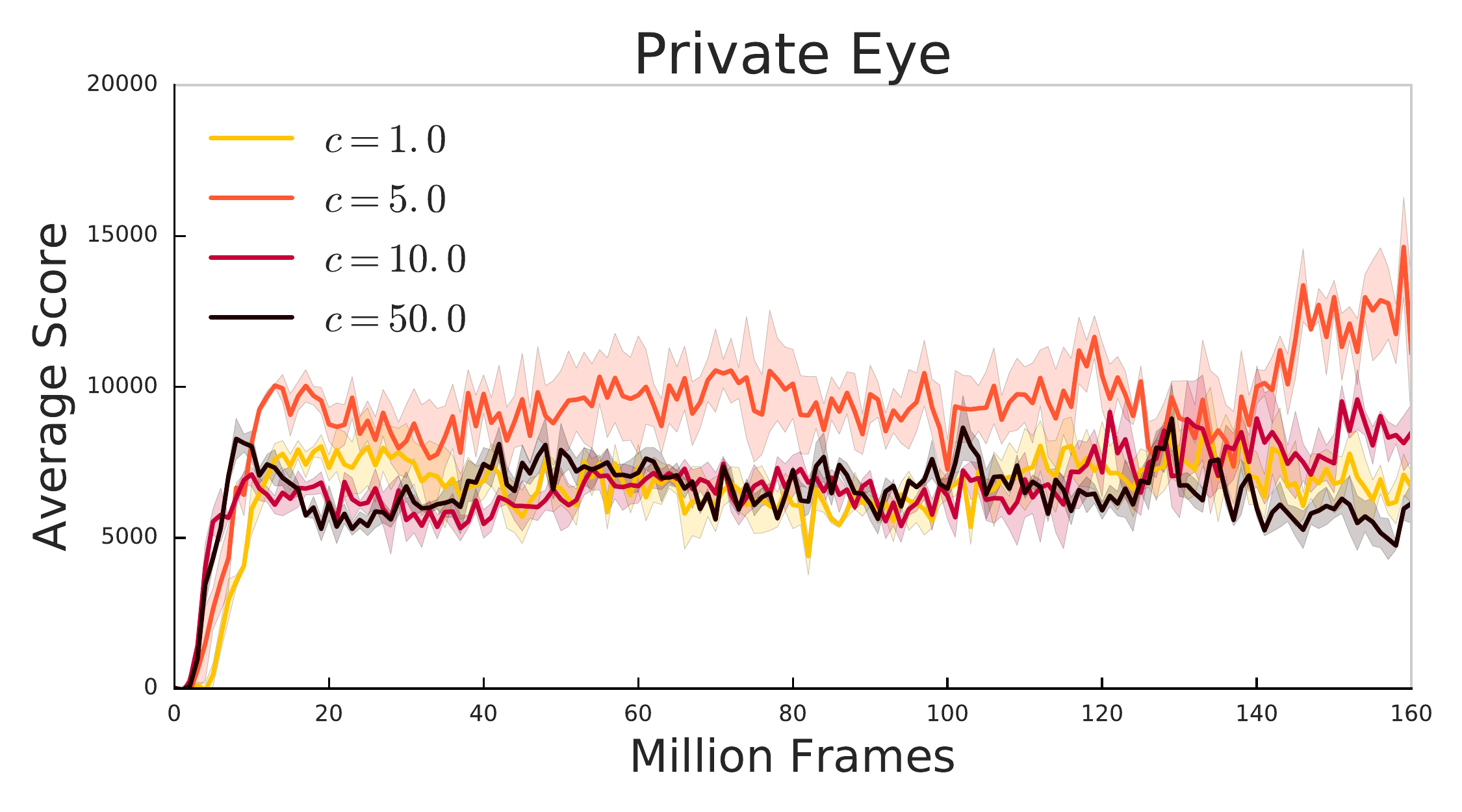}
}
\vspace{-1em}
\caption{DQN-PixelCNN trained from \textbf{intrinsic reward only} (3 seeds for each
configuration).}
\label{fig:im_only}
\end{figure}

Another way of creating an entirely curiosity-driven agent is to ignore
the environment reward altogether
and train based on the exploration reward only, see \figref{im_only}.
Remarkably, the curiosity signal alone is sufficient to train
a high-performing agent (measured by environment reward!).

It is worth noting that agents with
exploration bonus seem to `never stop exploring': for different seeds,
the agents make learning progress at very different times
during training, a qualitative difference to vanilla DQN.

\section{Conclusion}

We demonstrated the use of PixelCNN for exploration
and showed that its greater accuracy and expressiveness translate into
a more useful exploration bonus than that obtained from previous models.
While the current theory of pseudo-counts puts stringent requirements on the
density model, we have shown that PixelCNN can be used in a simpler and more general setup,
and can be trained completely online. It also proves to be
widely compatible with both value-function and policy-based RL algorithms.

In addition to pushing the state of the art on the
hardest exploration problems among the Atari 2600 games, PixelCNN
improves speed of learning and stability of baseline RL agents across
a wide range of games. The quality of its reward bonus is
evidenced by the fact that on sparse reward games, this signal alone
suffices to learn to achieve significant scores, creating a truly
intrinsically motivated agent.

Our analysis also reveals the importance of the Monte Carlo return
for effective exploration. The comparison with more sophisticated but fixed-horizon
multi-step methods shows that its significance lies both in faster learning
in the context of a useful but transient reward function, as well as
bridging reward gaps in environments where extrinsic and intrinsic
rewards are, or quickly become, extremely sparse.

\section*{Acknowledgements}
The authors thank Tom Schaul, Olivier Pietquin, Ian Osband, Sriram Srinivasan, Tejas Kulkarni,
Alex Graves, Charles Blundell, and Shimon Whiteson
for invaluable feedback on the ideas presented here,
and Audrunas Gruslys especially for providing the Reactor agent.

\bibliography{pixelcnn-counts}
\bibliographystyle{icml2017}

\clearpage
\newpage

\noappendix{
\appendix

\section{PixelCNN Hyper-parameters} \label{sec:hyperparams}

The PixelCNN model used in this paper is a lightweight variant of the Gated PixelCNN
introduced in \cite{oord2016pixelb}. It consists of a $7 \times 7$ masked convolution, followed by two residual blocks with $1 \times 1$ masked convolutions with 16 feature planes, and another $1 \times 1$
masked convolution producing $64$ features planes, which are mapped by a final
masked convolution to the output logits.
Inputs are $42 \times 42$ greyscale images, with pixel values quantized to $8$ bins.

The model is trained completely online, from the stream of Atari frames experienced
by an agent. Optimization is performed with the (uncentered) RMSProp optimizer \cite{tieleman2012lecture} with momentum $0.9$, decay $0.95$ and epsilon $10^{-4}$.

\section{Methodology} \label{sec:methodology}

Unless otherwise stated, all agent performance graphs in this paper show the agent's
training performance, measured as the undiscounted per-episode return, averaged over
1M environment frames per data point.

The algorithm-comparison graphs \figref{pcnn_cts_improv} and \figref{reactor_improv}
show the relative improvement of one algorithm over another in terms of area-under-the-curve (AUC). A comparison by maximum achieved score would yield
similar overall results, but underestimate the advantage in terms of learning speed
(sample efficiency) and stability that the intrinsically motivated
and MMC-based agents show over the baselines.

\section{Convolutional CTS} \label{sec:convcts}

\begin{figure*}
\center{
\includegraphics[width=2.22in]{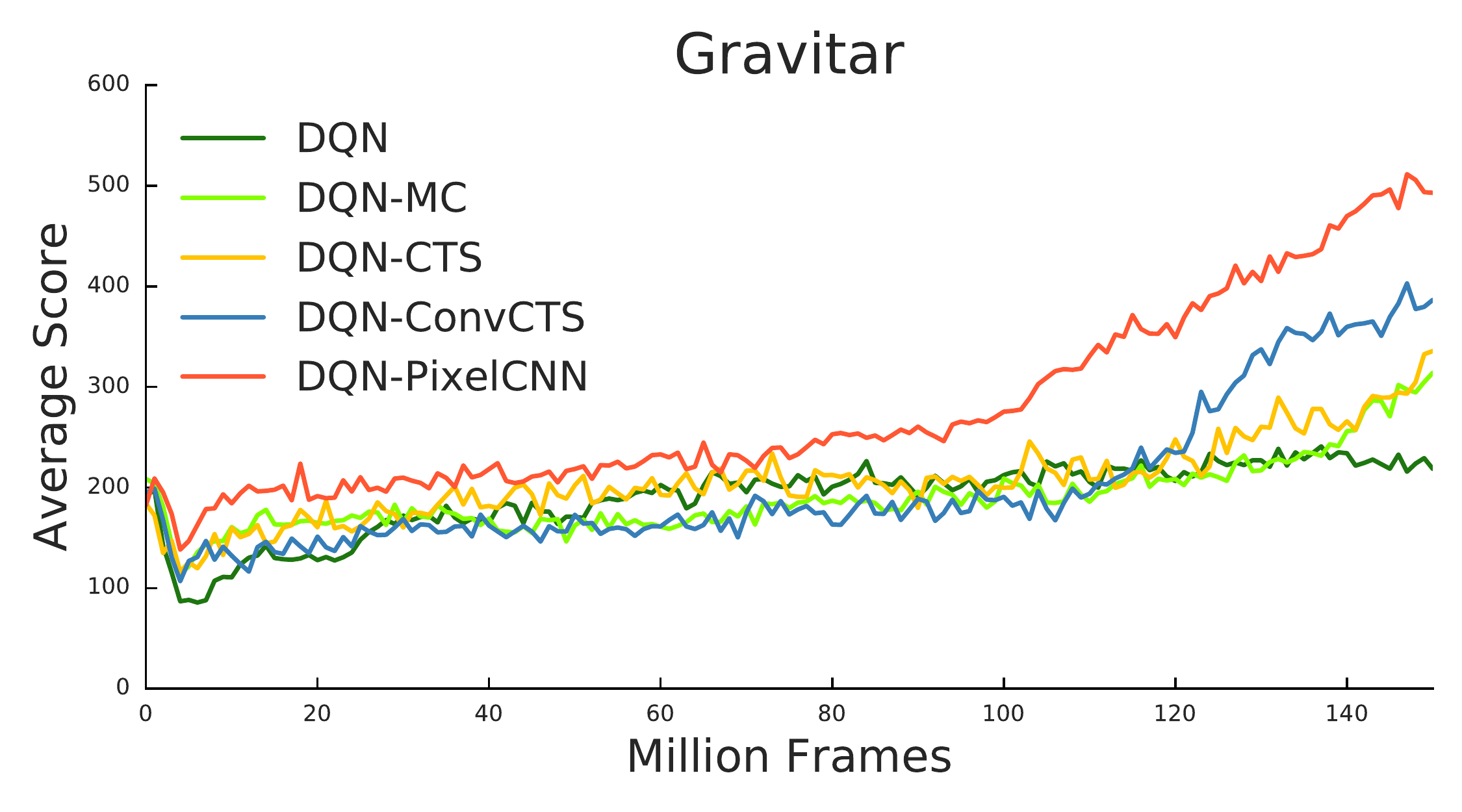}
\includegraphics[width=2.22in]{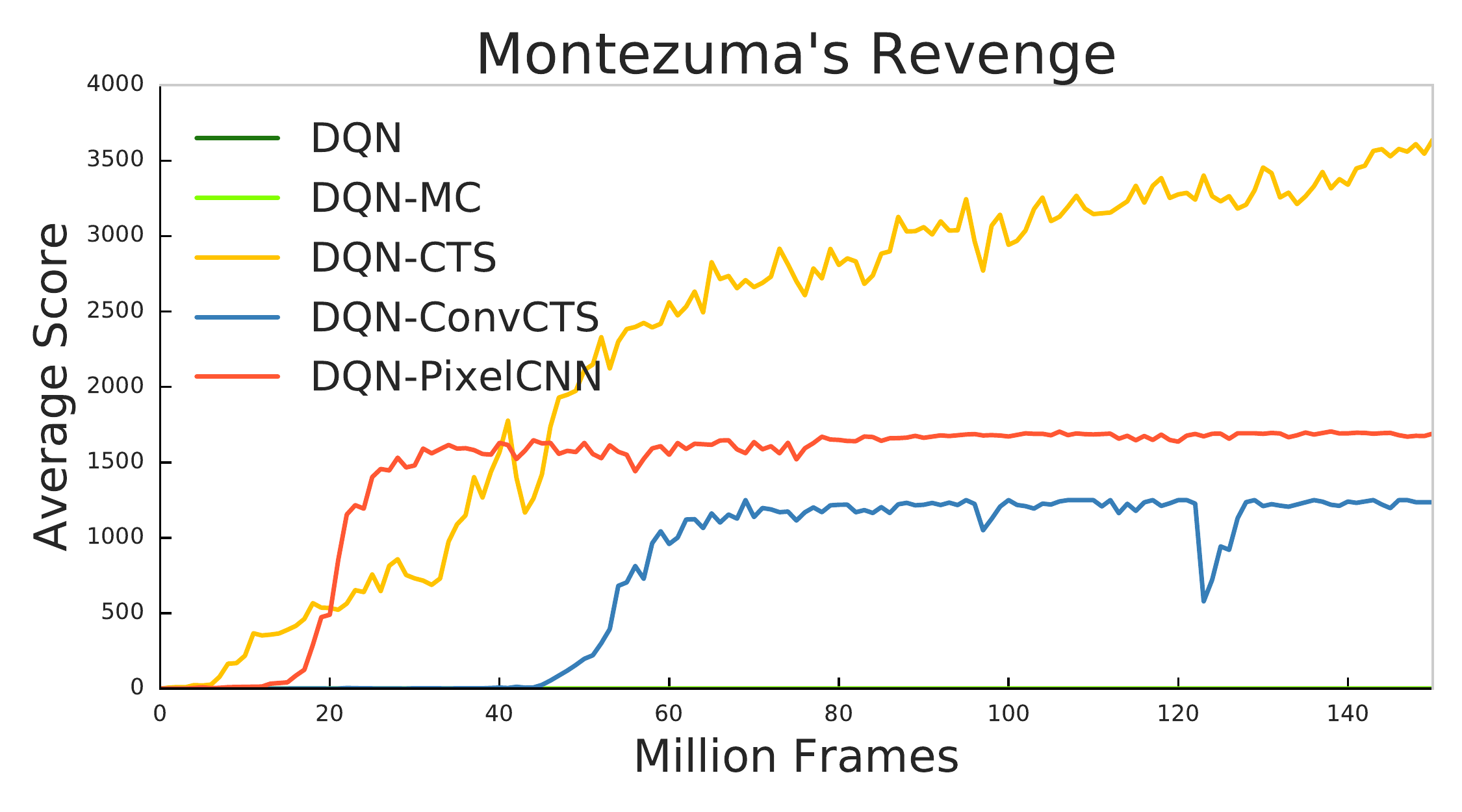}
\includegraphics[width=2.22in]{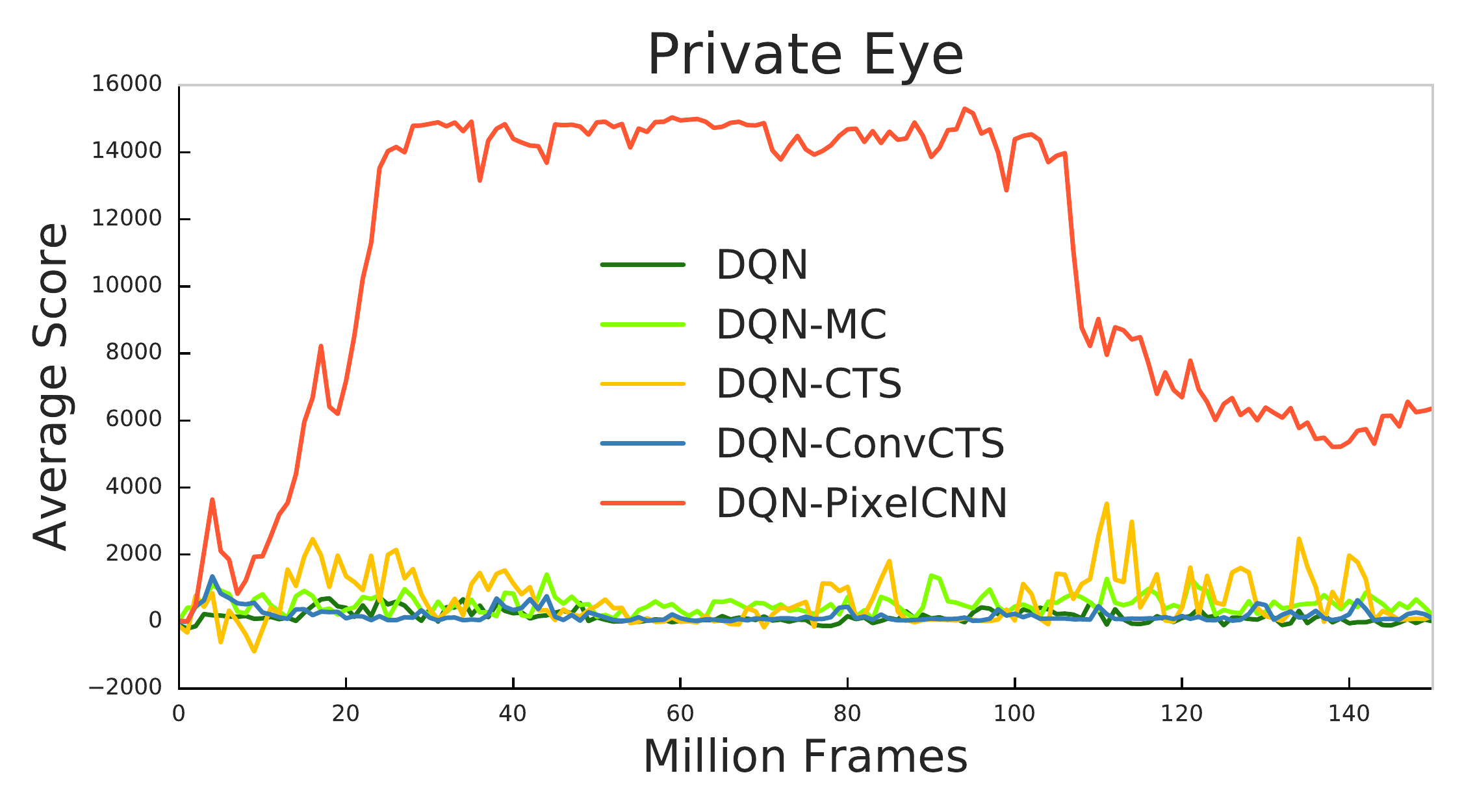}
\includegraphics[width=2.22in]{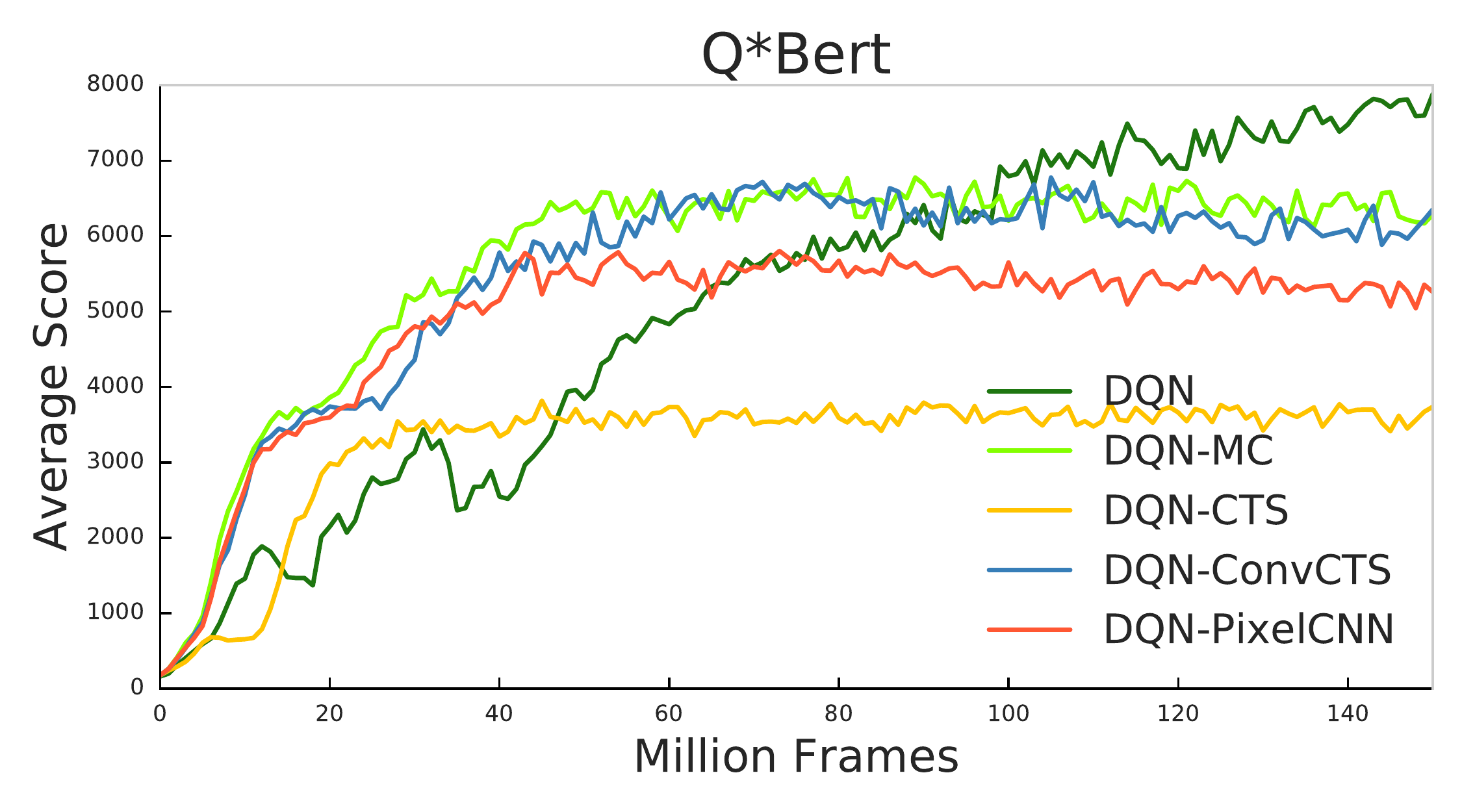}
\includegraphics[width=2.22in]{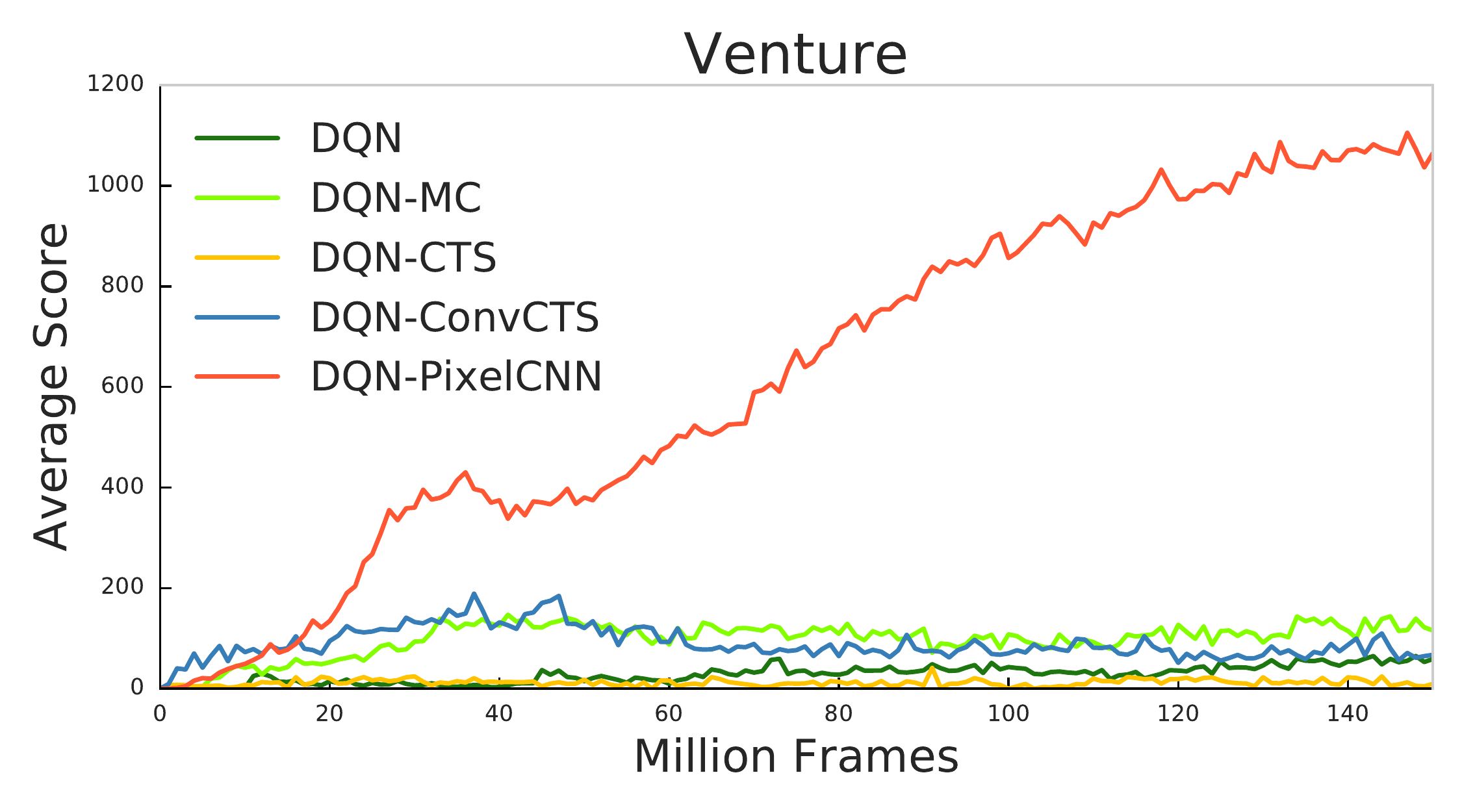}
\includegraphics[width=2.22in]{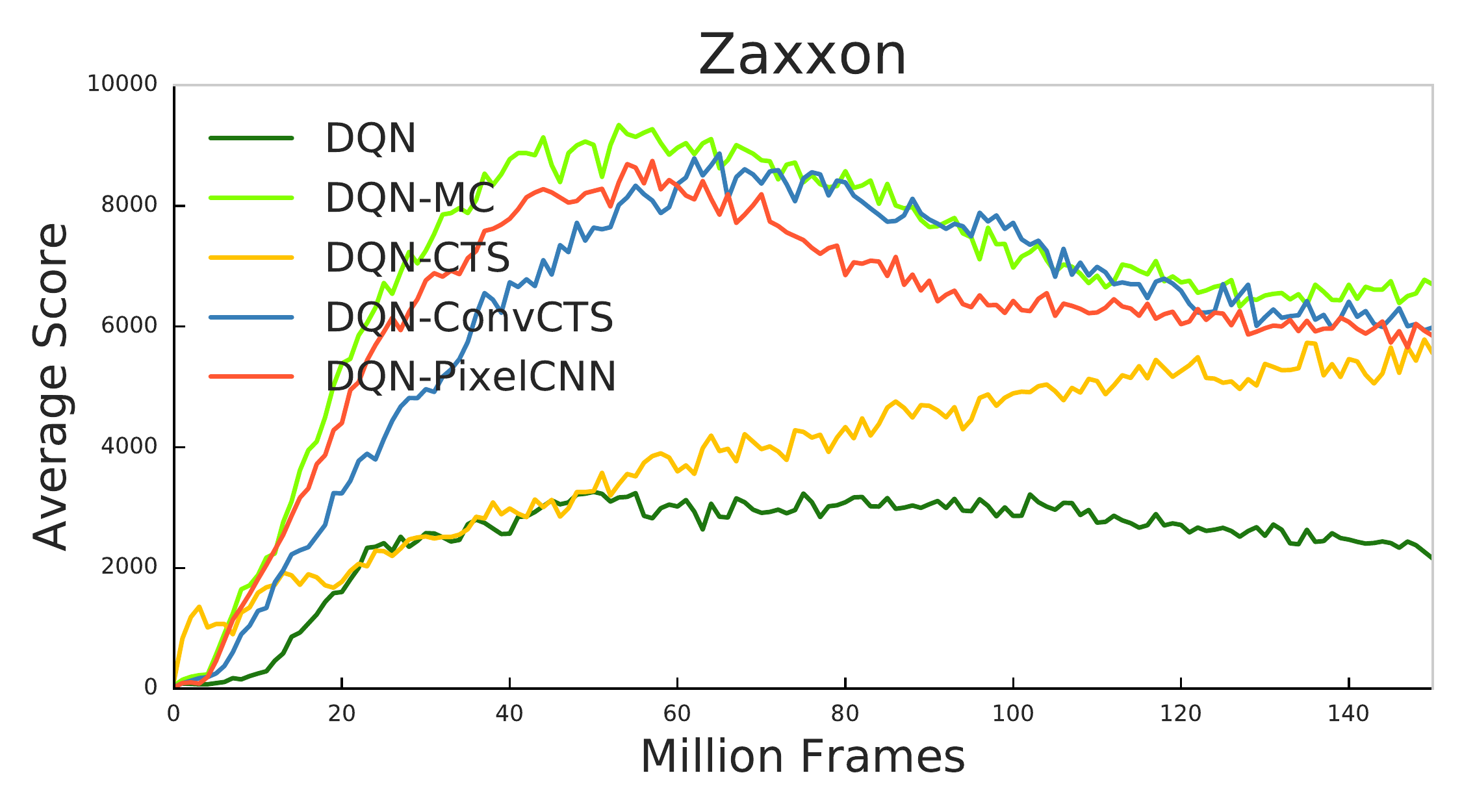}
}\caption{Comparison of DQN, DQN-CTS, DQN-ConvCTS and DQN-PixelCNN training performance.
\label{fig:convcts}}
\end{figure*}

In Section \ref{sec:atari} we have seen that DQN-PixelCNN outperforms DQN-CTS
in most of the 57 Atari games, by providing a more impactful exploration
bonus in hard exploration games, as well as a more graceful (less harmful)
one in games where the learning algorithm does not benefit from the
additional curiosity signal. One may wonder whether this improvement is due
to the generally more expressive and accurate density model PixelCNN,
or simply its convolutional nature, which gives it an advantage in generalization
and sample efficiency over a model that represents pixel probabilities in a completely
location-dependent way.

To answer this question, we developed a convolutional variant of the CTS model.
This model has a single set of parameters
conditioning a pixel's value on its predecessors shared across all pixel locations,
instead of the location-dependent parameters in the regular CTS.
In \figref{convcts} we contrast the performance of DQN, DQN-MC, DQN-CTS, DQN-ConvCTS and
DQN-PixelCNN on 6 example games.

We first consider dense reward games like \textsc{Q*Bert} and \textsc{Zaxxon},
where most improvement comes from the use of the MMC, and
the exploration bonus hurts performance. We find that in fact convolutional CTS
behaves fairly similarly to PixelCNN, leaving agent performance unaffected, whereas regular
CTS causes the agent to train more slowly or reach an earlier performance plateau.
On the sparse reward games (\textsc{Gravitar}, \textsc{Private Eye}, \textsc{Venture})
however, convolutional CTS shows to be as inferior to PixelCNN as the vanilla CTS
variant, failing to achieve the significant improvements over the baseline agents presented
in this paper.

We conclude that while the convolutional aspect plays a role in the 'softer' nature of the
PixelCNN model compared to its CTS counterpart, it alone is insufficient to
explain the massive exploration boost that the PixelCNN-derived reward provides
to the DQN agent. The more advanced model's accuracy advantage translates
into a more targeted and useful curiosity signal for the agent, which distinguishes
novel from well-explored states more clearly and allows for more effective exploration.

\section{The Hardest Exploration Games}\label{sec:taxonomy}

\begin{table*}
\center{
\scriptsize{
\begin{tabular}{|cc|c||c|c|}
\hline
\multicolumn{3}{c}{\normalsize{Easy Exploration}} & \multicolumn{2}{c}{\normalsize{Hard Exploration}} \\
\hline
\hline
\multicolumn{2}{c}{\small{Human-Optimal}} & \multicolumn{1}{c}{\small{Score Exploit}} & \multicolumn{1}{c}{\small{Dense Reward}} & \multicolumn{1}{c}{\small{Sparse Reward}} \\
\hline
\gamename{Assault } &\gamename{ Asterix } &\gamename{ Beam Rider } &\gamename{ Alien } &\gamename{ Freeway } \\
\hline
\gamename{Asteroids } &\gamename{ Atlantis } &\gamename{ Kangaroo } &\gamename{ Amidar } &\gamename{ Gravitar } \\
\hline
\gamename{Battle Zone } &\gamename{ Berzerk } &\gamename{ Krull } &\gamename{ Bank Heist } &\gamename{ Montezuma's Revenge } \\
\hline
\gamename{Bowling } &\gamename{ Boxing } &\gamename{ Kung-fu Master } &\gamename{ Frostbite } &\gamename{ Pitfall! } \\
\hline
\gamename{Breakout } &\gamename{ Centipede } &\gamename{ Road Runner } &\gamename{ H.E.R.O. } &\gamename{ Private Eye } \\
\hline
\gamename{Chopper Cmd } &\gamename{ Crazy Climber } &\gamename{ Seaquest } &\gamename{ Ms. Pac-Man } &\gamename{ Solaris } \\
\hline
\gamename{Defender } &\gamename{ Demon Attack } &\gamename{ Up n Down } & \gamename{Q*Bert} & \gamename{Venture} \\
\hline
\gamename{Double Dunk } &\gamename{ Enduro } & \gamename{ Tutankham } &\gamename{ Surround } & \\
\hline
\gamename{Fishing Derby } &\gamename{ Gopher } &  &\gamename{ Wizard of Wor } & \\
\hline
\gamename{Ice Hockey } &\gamename{ James Bond } &  &\gamename{ Zaxxon } & \\
\hline
\gamename{Name this Game } &\gamename{ Phoenix } &  & & \\
\hline
\gamename{Pong } &\gamename{ River Raid } & &  & \\
\hline
\gamename{Robotank } &\gamename{ Skiing } &  &  & \\
\hline
\gamename{Space Invaders } &\gamename{ Stargunner } &  &  & \\
\hline
\end{tabular}
}}
\caption{A rough taxonomy of Atari 2600 games according to their exploration difficulty.\label{table:taxonomy}}
\end{table*}

\begin{table*}
\center{
\small
\begin{tabular}{|r|r|r|r||r|r|}
\hline
& DQN & A3C-CTS & Prior. Duel & DQN-CTS & DQN-PixelCNN \\
\hline
\textsc{Freeway}   & 30.8 & 30.48 & \textbf{33.0} & 31.7 & 31.7 \\
\hline
\textsc{Gravitar}   & 473.0 & 238.68 & 238.0 & 498.3 &\textbf{859.1} \\
\hline
\textsc{Montezuma's Revenge} & 0.0 & 273.70 & 0.0 & \textbf{3705.5} & 2514.3 \\
\hline
\textsc{Pitfall!} & -286.1 & -259.09 & \textbf{0.0} & \textbf{0.0} &\textbf{0.0} \\
\hline
\textsc{Private Eye} &  146.7 & 99.32 & 206.0 & 8358.7 &\textbf{15806.5} \\
\hline
\textsc{Solaris} & 3,482.8 & 2270.15 & 133.4 & 2863.6 &\textbf{5501.5} \\
\hline
\textsc{Venture} & 163.0 & 0.00 & 48.0 & 82.2 &\textbf{1356.25}\\
\hline
\end{tabular}
}
\caption{Comparison with previously published results on hard exploration, sparse reward games. The compared agents are
DQN \cite{mnih2015human}, A3C-CTS (``A3C+'' in \cite{bellemare16cts}),
Prioritized Dueling DQN \cite{wang2016dueling},
and the basic versions of DQN-CTS and DQN-PixelCNN from Section \ref{sec:atari}.
For our agents we report the maximum scores achieved over 150M frames of training, averaged over 3 seeds.}
\label{table:hard_games_results}
\end{table*}

Table \ref{table:taxonomy} reproduces \citet{bellemare16cts}'s taxonomy of games available through the ALE according to their exploration difficulty.
``Human-Optimal'' refers to games where DQN-like agents achieve human-level or higher performance;
``Score Exploit'' refers to games where agents find ways to achieve superhuman scores, without necessarily playing the game as a human would.
``Sparse'' and ``Dense'' rewards are qualitative descriptors of the game's reward structure. See the original source for additional details.

Table \ref{table:hard_games_results} compares previously published results on the 7 hard exploration, sparse reward Atari 2600 games with
results obtained by DQN-CTS and DQN-PixelCNN.

\begin{figure*}
\begin{center}
\includegraphics[width=0.85\textwidth]{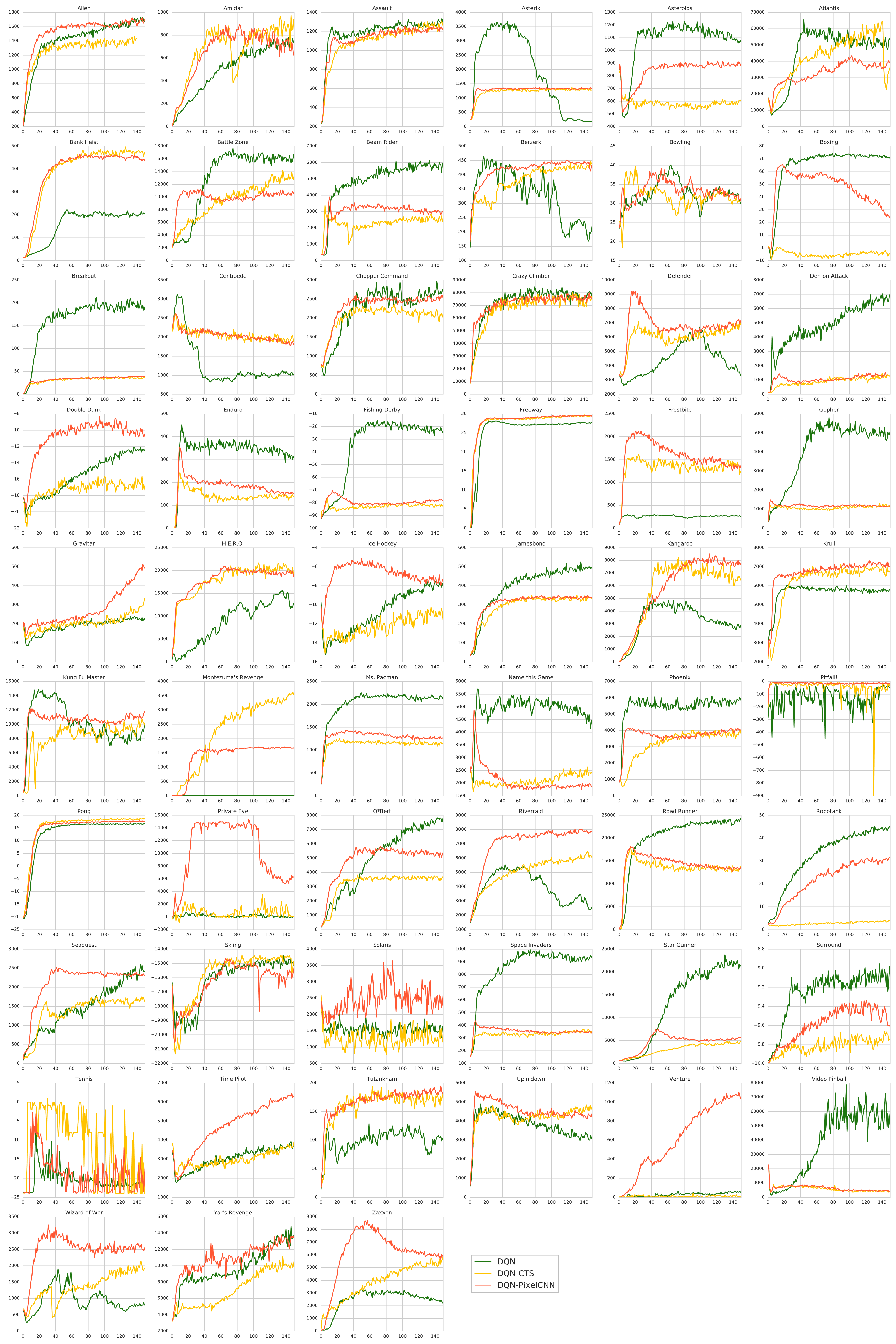}
\caption{Training curves of DQN, DQN-CTS and DQN-PixelCNN across all 57 Atari games.}
\label{fig:pcnn_cts_full}
\end{center}
\end{figure*}

\begin{figure*}
\center{
\includegraphics[width=0.85\textwidth]{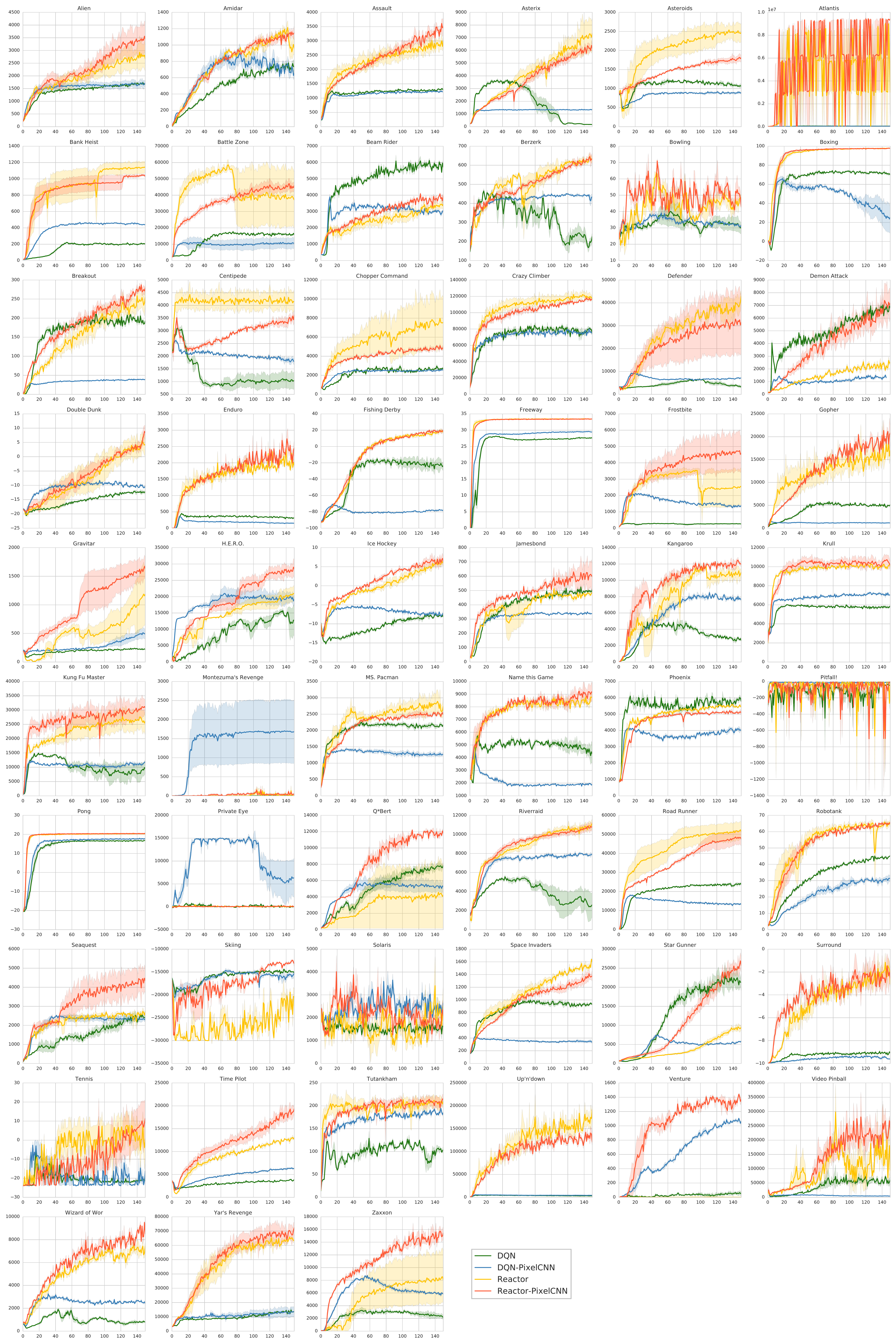}
}\caption{Training curves of DQN, DQN-PixelCNN, Reactor and Reactor-PixelCNN across all 57 Atari games.
\label{fig:reactor_full}}
\end{figure*}

\begin{figure*}
\center{
\includegraphics[width=0.85\textwidth]{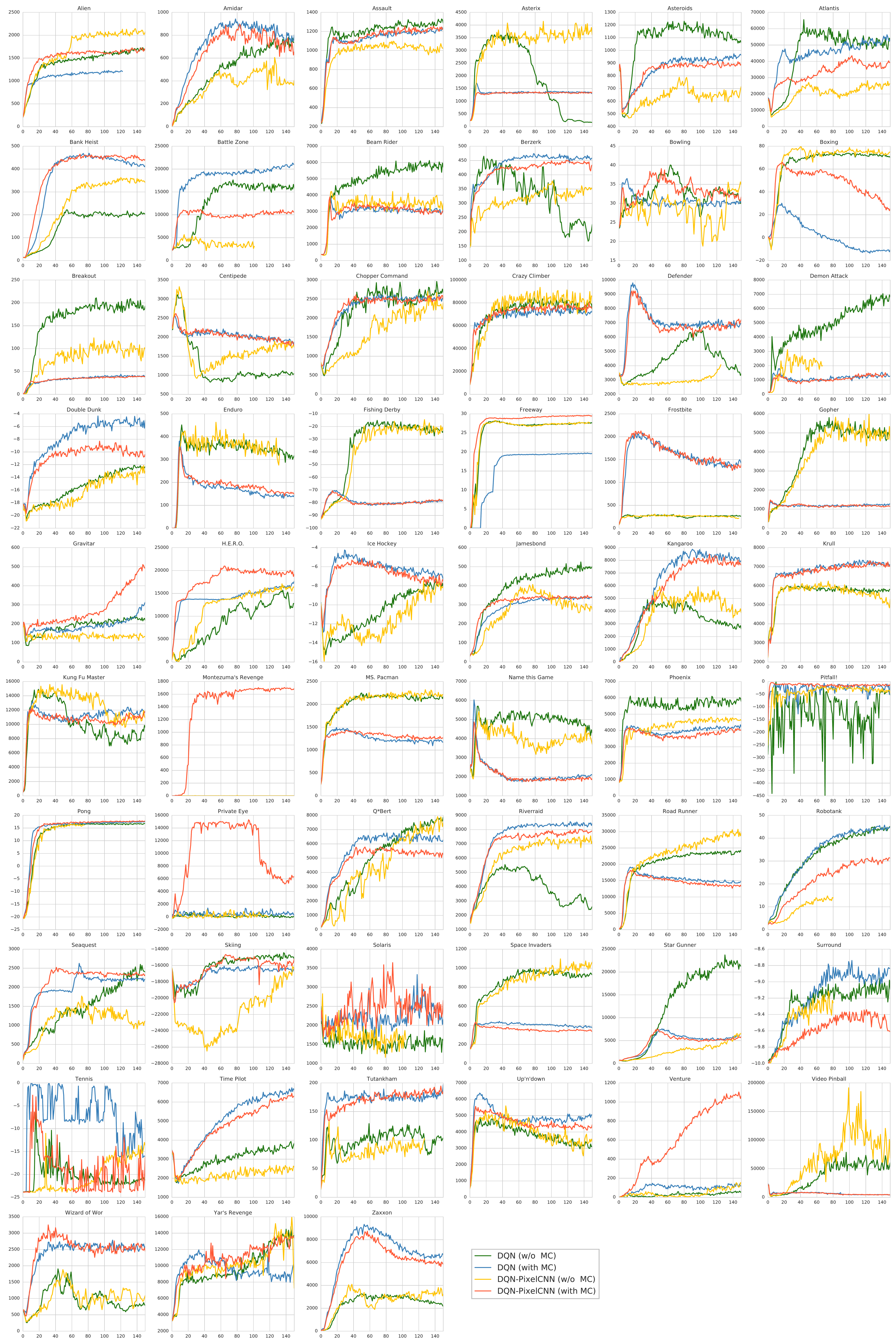}
}\caption{Training curves of DQN and DQN-PixelCNN, each with and without MMC, across all 57 Atari games.
\label{fig:mc_full}}
\end{figure*}

} 

\end{document}